\pdfoutput=1

\documentclass[11pt]{article}

\usepackage{EMNLP2022}

\usepackage{times}
\usepackage{latexsym}
\usepackage{hyperref}
\usepackage{amsmath}
\usepackage[normalem]{ulem}
\usepackage{graphicx}
\usepackage{enumitem}  
\usepackage{booktabs}
\usepackage{multirow}
\usepackage{amssymb}
\usepackage{pifont}
\usepackage{lscape}

\usepackage{colortbl}
\definecolor{aliceblue}{rgb}{0.94, 0.97, 1.0}
\definecolor{chromeyellow}{rgb}{1.0, 0.65, 0.0}
\definecolor{ceruleanblue}{rgb}{0.16, 0.32, 0.75}
\definecolor{cinnabar}{rgb}{0.89, 0.26, 0.2}
\definecolor{asparagus}{rgb}{0.53, 0.66, 0.42}
\definecolor{bittersweet}{rgb}{1.0, 0.44, 0.37}
\definecolor{blue(pigment)}{rgb}{0.2, 0.2, 0.6}

\usepackage{gb4e}
\noautomath
\usepackage{tikz} 				
\usepackage{tikz-qtree}	
	\usetikzlibrary{fit}			
	\tikzset{every tree node/.style={align=center, anchor=north}}
\usepackage{graphicx}

\usepackage{array}
	\newcolumntype{K}[1]{>{\centering\arraybackslash}p{#1}}
\usepackage[normalem]{ulem}
\usepackage{CJKutf8}

\usepackage[T1]{fontenc}

\usepackage[utf8]{inputenc}

\usepackage{microtype}

\newcommand{\titlestr}{\dataset: \textsc{S}ino \textsc{Ling}uistic Evaluation of Large Language Models}

\title{\titlestr}

\author{Yixiao Song$^{\diamondsuit}$ \quad Kalpesh Krishna$^{\spadesuit}$ \quad Rajesh Bhatt$^\diamondsuit$ \quad Mohit Iyyer$^\spadesuit$ \\\\
$^\diamondsuit$Department of Linguistics, UMass Amherst \\ $^\spadesuit$Manning College of Information and Computer Sciences, UMass Amherst \\ \texttt{\{yixiaosong,bhatt\}@umass.edu} \\ \texttt{\{kalpesh,miyyer\}@cs.umass.edu} }

\newcommand{\dataset}{\textsc{Sling}}

\newcommand{\namedref}[2]{\hyperref[#2]{#1~\ref*{#2}}}

\newcommand{\sectionref}[1]{\namedref{Section}{#1}}
\newcommand{\tableref}[1]{\namedref{Table}{#1}}
\newcommand{\figureref}[1]{\namedref{Figure}{#1}}
\newcommand{\appendixref}[1]{\namedref{Appendix}{#1}}

\newcommand{\yscomment}[1]{\textcolor{purple}{\bf \small [#1 --YS]}}

\newcommand{\model}[1]{\texttt{#1}}

\begin{document}
\maketitle

\begin{abstract}
To understand what kinds of linguistic knowledge are encoded by pretrained Chinese language models (LMs), we introduce the benchmark of \textsc{S}ino \textsc{Ling}uistics (\dataset), which consists of $38$K minimal sentence pairs in Mandarin Chinese grouped into 9 high-level linguistic phenomena. Each pair demonstrates the acceptability contrast of a specific syntactic or semantic phenomenon (e.g., The keys \textit{are} lost \textit{vs.} The keys \textit{is} lost), and an LM should assign lower perplexity to the acceptable sentence. In contrast to the CLiMP dataset~\citep{climp}, which also contains Chinese minimal pairs and was created by translating the vocabulary of the English BLiMP dataset, the minimal pairs in \dataset\ are derived primarily by applying syntactic and lexical transformations to naturally-occurring, linguist-annotated sentences from the Chinese Treebank 9.0, thus addressing severe issues in CLiMP's data generation process.
We test 18 publicly available pretrained monolingual (e.g., \model{BERT-base-zh}, \model{CPM}) and multi-lingual (e.g., \model{mT5}, \model{XLM}) language models on \dataset. Our experiments show that the average accuracy for LMs is far below human performance ($69.7$\% vs. $97.1$\%), while \model{BERT-base-zh} achieves the highest accuracy ($84.8$\%) of all tested LMs, even much larger ones. Additionally, we find that most LMs have a strong gender and number (singular/plural) bias, and they perform better on local phenomena than hierarchical ones.\footnote{The \dataset\ data and code can be found \url{https://github.com/Yixiao-Song/SLING_Data_Code}.} 

\end{abstract}
\section{Introduction}

While large-scale pretrained language models (LMs) have achieved considerable downstream success~\citep[][a.o.]{devlin2018bert,xue2020mt5,brown2020language}, it remains challenging to evaluate how much linguistic knowledge they have acquired.  One approach is to design \emph{minimal pairs} consisting of two sentences that differ only in a critical word or phrase, which renders only one of the sentences acceptable (e.g., \textit{The keys are lost} vs. \textit{The keys is lost}). If an LM is sensitive to the phenomenon exemplified by the minimal pair (in this case, plurality), it should assign a lower perplexity to the acceptable sentence. 
This methodology can be used to test an LM's understanding of a wide range of linguistic phenomena; for example, the BLiMP dataset~\citep{blimp} contains 67K minimal pairs automatically generated via manually-constructed grammars that span 12 high-level English phenomena.

\begin{figure}[t]
    \centering
    \includegraphics[width=0.48\textwidth]{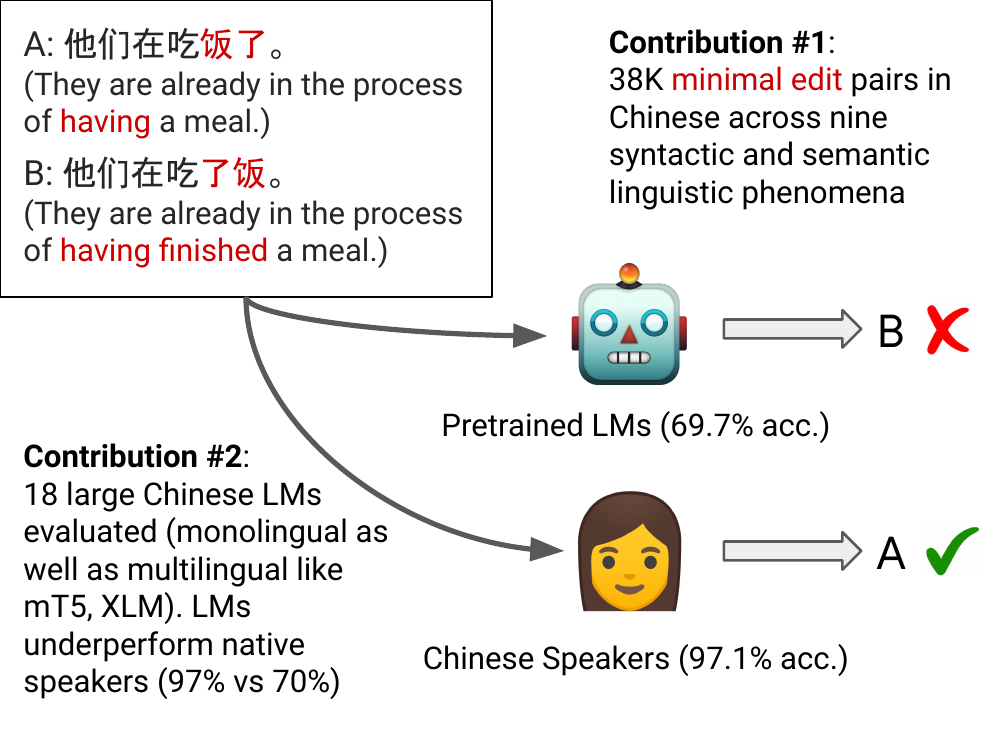}
    \caption{An illustration of the \dataset~dataset. The \textit{A} sentence is acceptable but \textit{B}, a minimal edit counterpart of \textit{A}, is not. LMs see one sentence at a time and are expected to assign a lower (pseudo-)perplexity to the acceptable sentence. Overall, LMs underperform Chinese native speakers on \dataset~(97\% vs 70\%), making it an exciting benchmark for future Chinese LM research.}
    \label{fig:sling-overview}
    \vspace{-0.15in}
\end{figure}

Can we create similar datasets to study linguistic phenomena in a different language, such as Chinese? As a first step in this direction,~\citet{climp} introduce CLiMP, a Chinese dataset of minimal pairs. However, we identify two major issues with CLiMP's construction process: (1) its vocabulary is translated from BLiMP's vocabulary, which due to morphological differences between English and Chinese (e.g., the latter lacks numeral or verbal inflections) results in a large number of unintelligible sentences; and (2) the grammatical templates for several phenomena (anaphor agreement, classifier-noun agreement, and filler-gap dependencies) are inadequately designed, which along with the vocabulary issue results in minimal pairs that do not have any clear contrast.\footnote{Note that although~\citet{climp} report a high human accuracy of $97.1$\% on CLiMP, this number is calculated using majority vote of 16 annotators, and the inter-annotator agreement is not reported.}



To address these issues, we introduce \dataset~(\textsc{s}ino \textsc{ling}uistics benchmark), a dataset of $38$K minimal pairs to study nine Chinese linguistic phenomena, many of which are unique to the Chinese language. Instead of translating BLiMP, we construct \dataset~primarily using the Chinese Treebank 9.0 \citep{ctb9}, which was annotated by trained linguists (see \tableref{tab:comparisonclimpsling} for a comparison). We extract subtrees from human-validated constituency parses in this treebank and then carefully edit them using manually-designed linguistic templates to create minimal pairs. \dataset\ does not suffer from the issues we found in CLiMP, and it additionally includes semantic as well as syntactic phenomena, seven of which are not found in CLiMP. A human validation of \dataset\ with 16 native speakers confirms that its minimal pairs unambiguously show the acceptability contrast across all phenomena, yielding an \emph{almost perfect} inter-annotator agreement (Fleiss' $\kappa$ = $0.88$).

We evaluate a total of 18 publicly-available pretrained LMs on \dataset, including monolingual Chinese (e.g., \model{bert-base-chinese}, \model{PanGu-$\alpha$}) and multilingual models (e.g., \model{mT5}, \model{XLM-R}).  Our results reveal that: (1) no LM consistently outperforms others on \dataset; (2) larger LMs do not necessarily outperform smaller ones; (3) monolingual Chinese LMs generally perform better than multilingual ones; and (4) humans significantly outperform all LMs ($97.1$\% vs $69.7$\% average across LMs). We observe that the ranking of models on CLiMP differs from that on \dataset: for example, \model{bert-chinese-base} is the best-performing model on \dataset~(average accuracy $84.8$\%), while \model{chinese-pert-base} performs best on CLiMP ($81.2$\%). This result is due in part to the issues in CLiMP's construction process, as well as the different phenomena that we test in \dataset. Additionally, \dataset\ is more discriminative than CLiMP (i.e., LMs vary more across the phenomena in terms of accuracy), which makes it more useful as a diagnostic benchmark especially given the large gap between human and model performance. 




\begin{table}[!t]
\centering
\resizebox{\linewidth}{!}{%
\begin{tabular}{@{}lll@{}}
\toprule
                               & \bf CLiMP                  & \bf \dataset                \\
\midrule
vocab. source                  & BLiMP's vocab. translated & Chinese Treebank 9.0 \\\midrule
\multirow{3}{*}{vocab. size} & actual 1272 types        &    \\
                             &  (w/ 230 proper names)   &  11988 types \\
                               & (claimed 3456)         &       \\\midrule
\multirow{2}{*}{grammar}     & 9 syntax phenomena                & 3 semantics + 6 syntax \\
& (16 paradigms) & (5 syntax differ from CLiMP)\\
                               &          & (38 paradigms)       \\\midrule
\multirow{4}{*}{evaluated LMs} & monolingual only      & 10 mono- \& 8 multilingual\\
                               & 1 bert-base-chinese    & 1 LSTM               \\
                               & 3 LSTM                 & 3 Causal LMs         \\
                               & 2 5-gram               & 14 Masked LMs         \\\bottomrule
\end{tabular}}
\caption{An comparison between CLiMP~\citep{climp} and \dataset. \dataset\ is created with a natural and diverse vocabulary, covers new semantic and syntactic Chinese linguistic phenomena, and is evaluated on large pretrained LMs, including multilingual models like mT5.\footnote{\yscomment{The size of vocabulary is uncertain because the definition of \textit{word} in Chinese is unclear \citep[Section 1]{xia2000segmentation}. \appendixref{appen.vocab_size} reports the ngram counts} of CLiMP and \dataset.}}
\label{tab:comparisonclimpsling}
\end{table}
\section{Evaluating Chinese LMs with Minimal Pairs: CLiMP and Its Shortcomings}


Using minimal pairs to detect a function of a single element (e.g., phoneme, affix, or word) is a common practice in linguistics. In \figureref{fig:sling-overview}, by changing the position of \begin{CJK*}{UTF8}{gbsn}\small 了\end{CJK*}, sentence \textit{A} is transformed into the ungrammatical sentence \textit{B}, and we know how the two aspect markers \begin{CJK*}{UTF8}{gbsn}\small
在\end{CJK*} and \begin{CJK*}{UTF8}{gbsn}\small 了\end{CJK*} interacts. In this paper, following BLiMP and CLiMP, we call each major grammatical category a \textit{phenomenon}, and minimal pair types within each phenomenon \emph{paradigms}. 
The \textit{A} and \textit{B} sentences in \figureref{fig:sling-overview} form a minimal pair of a paradigm in the \emph{aspect} phenomenon of \dataset.\footnote{More examples of minimal pairs can be found in \appendixref{appen.linguistic.phenomena}.} 

\citet{climp} created CLiMP to evaluate 9 Chinese syntactic phenomena with 16 paradigms. However, the dataset suffers from two major issues: (1) faulty minimal pair generation templates and (2) its translated vocabulary. In this section, we discuss the issues in detail and show why they hamper CLiMP's utility as a diagnostic dataset for LMs. 

\paragraph{CLiMP's minimal pairs often do not show the desired acceptability contrast.} This problem is especially prominent in the \textit{ba} construction, binding/anaphor, and filler-gap dependency phenomena, on which~\citet{climp} conclude that LMs perform poorly. The templates used to generate data for these phenomena are the primary cause of these errors, as we show below.

\paragraph{\textit{ba} construction:} Many minimal pairs associated with this construction do not exhibit the acceptability contrast.\footnote{The \textit{ba} construction is a way to move the object from its base position (after a verb) to the position before the verb. The construction expresses the meaning of \textit{settlement} and focuses on what is happening to the object.} We examine the first 50 minimal pairs 
of this phenomenon in CLiMP and discover that 6 pairs actually have the wrong acceptability label:

    \begin{table}[!h]
    \centering
    \resizebox{\linewidth}{!}{%
    \begin{tabular}{@{}lcc@{}}
    \toprule
    Sentences & CLiMP & Actual\\
    \midrule
    \begin{CJK*}{UTF8}{gbsn}\footnotesize 报告把大学转移了。\end{CJK*} The report relocated the university. & \ding{51} & \ding{55} \\
    \begin{CJK*}{UTF8}{gbsn}\footnotesize 报告被大学转移了。\end{CJK*} The report was relocated by the university.& \ding{55} & \ding{51} \\
    \bottomrule
    \end{tabular}}
    \end{table}
  
\noindent at least 9 minimal pairs contain two acceptable sentences:
  
    \begin{table}[!h]
    \centering
    \resizebox{\linewidth}{!}{%
    \begin{tabular}{@{}lcc@{}}
    \toprule
    Sentences & CLiMP & Actual\\
    \midrule
    \begin{CJK*}{UTF8}{gbsn}\footnotesize 吴宇涛把图书馆调查了。\end{CJK*} Wu investigated the library. & \ding{51} & \ding{51} \\
    \begin{CJK*}{UTF8}{gbsn}\footnotesize 吴宇涛被图书馆调查了。\end{CJK*} Wu was investigated by the library. & \ding{55} & \ding{51} \\
    \bottomrule
    \end{tabular}}
    \end{table}
    
\noindent and 4 pairs are unintelligible or nonsensical:

    \begin{table}[!h]
    \centering
    \resizebox{\linewidth}{!}{%
    \begin{tabular}{@{}lcc@{}}
    \toprule
    Sentences & CLiMP & Actual\\
    \midrule
    \begin{CJK*}{UTF8}{gbsn}\small 王萍把嘴举了\end{CJK*} Wang lifted a mouth. & \ding{51} & \ding{55} \\
    \begin{CJK*}{UTF8}{gbsn}\small 王萍被嘴举了\end{CJK*} Wang was lifted by a mouth. & \ding{55} & \ding{55} \\
    \bottomrule
    \end{tabular}}
    \end{table}    
    
    The primary reason for the low quality of these pairs is that CLiMP does not carefully control the source of unacceptability \citep{abrusan2019semantic}, which we discuss further in the Limitations section. Specific to the \textit{ba} construction, CLiMP does not include essential information about thematic relations\footnote{A thematic relation represents the semantic relation that a noun phrase bears with respect to an event denoted by a verb. For example, the thematic relation that \textit{John} holds to the verb \textit{eat} in \textit{John eats an apple.}is that of agent, which means \textit{John} is the agent of an apple eating event.} in the vocabulary. 
    Another contributing factor is the small size of the CLiMP vocabulary, which is translated from that of BLiMP despite many annotated features of BLiMP not applying to Chinese (e.g., number features, verb forms, or cases). For example, the English verb \textit{buy} has six forms in BLiMP, listed in \tableref{tab:climprepetitive}, which differ from each other in seven verb-related features. These inflections are useful in English for distinguishing sentence acceptability in several BLiMP phenomena (e.g., \emph{Passive}, \emph{Irregular Forms}, and \emph{Subject-Verb Agreement}); however, they do not apply to Chinese because the language lacks inflection, and thus they cannot help construct Chinese paradigms. In Chinese, the same forms can be represented and built based on the three words shown in bold: \textit{mai} (buy), \textit{(zheng) zai} (progressive marker), and \textit{le} (perfective marker). They do not need to be redundantly listed in the vocabulary. After removing the redundant word types, CLiMP's vocabulary size is 1,272 (including 230 proper names), not 3,456 as \citet{climp} report. This lack of diversity in the vocabulary contributes to the generation of nonsensical sentences using their minimal pair templates.

\begin{table}[!h]
\centering
\resizebox{0.75\linewidth}{!}{%
\begin{tabular}{@{}lll@{}}
\toprule
Chinese & English & Features\\
\midrule
\textbf{mai}           & buy  & bare \\
\textbf{zheng zai} mai & buying & ing\\
mai \textbf{le}        & bought & finite, past\\
mai le        & bought & en \\
mai           & buy    & finite, pres\\
mai           & buys   & finite, pres, 3sg \\ 
\bottomrule
\end{tabular}}
\caption{An example of the repetitive word types in CLiMP's vocabulary (\emph{mai} here). ing = progressive, en = participle, pres = present, 3sg = third person singular.}
\label{tab:climprepetitive}
\end{table}

  \paragraph{Binding and anaphor paradigms:} These two paradigms test whether the gender feature of the {\color{chromeyellow}object} anaphor agrees with that of the {\color{ceruleanblue}subject}. Issues in the binding and anaphor paradigms stem from the fact that CLiMP uses {\color{ceruleanblue}proper names}, which were added to CLiMP's vocabulary in addition to the one translated from BLiMP. However, Chinese proper names do not always unambiguously show gender.  If the gender of the {\color{ceruleanblue}subject} is ambiguous as in (\ref{ex.ambiguous.proper.name}) where {\color{ceruleanblue}Ye Zi} can be either gender (similarily for \textit{Alex} in English), the performance of the LMs is not representative of whether they know the function of the reflexive anaphor, which is exactly what the binding and anaphor paradigms want to test.
  
  \begin{exe}
    \ex\label{ex.ambiguous.proper.name}
        \begin{CJK*}{UTF8}{gbsn}\small {\color{ceruleanblue}叶梓}逃离了{\color{chromeyellow}他 / 她自己}。\end{CJK*}
    
        {\color{ceruleanblue}Ye Zi} escaped from {\color{chromeyellow}him- / herself}.
  \end{exe}

 \begin{figure*}[t]
    \centering
    \includegraphics[width=0.98\textwidth]{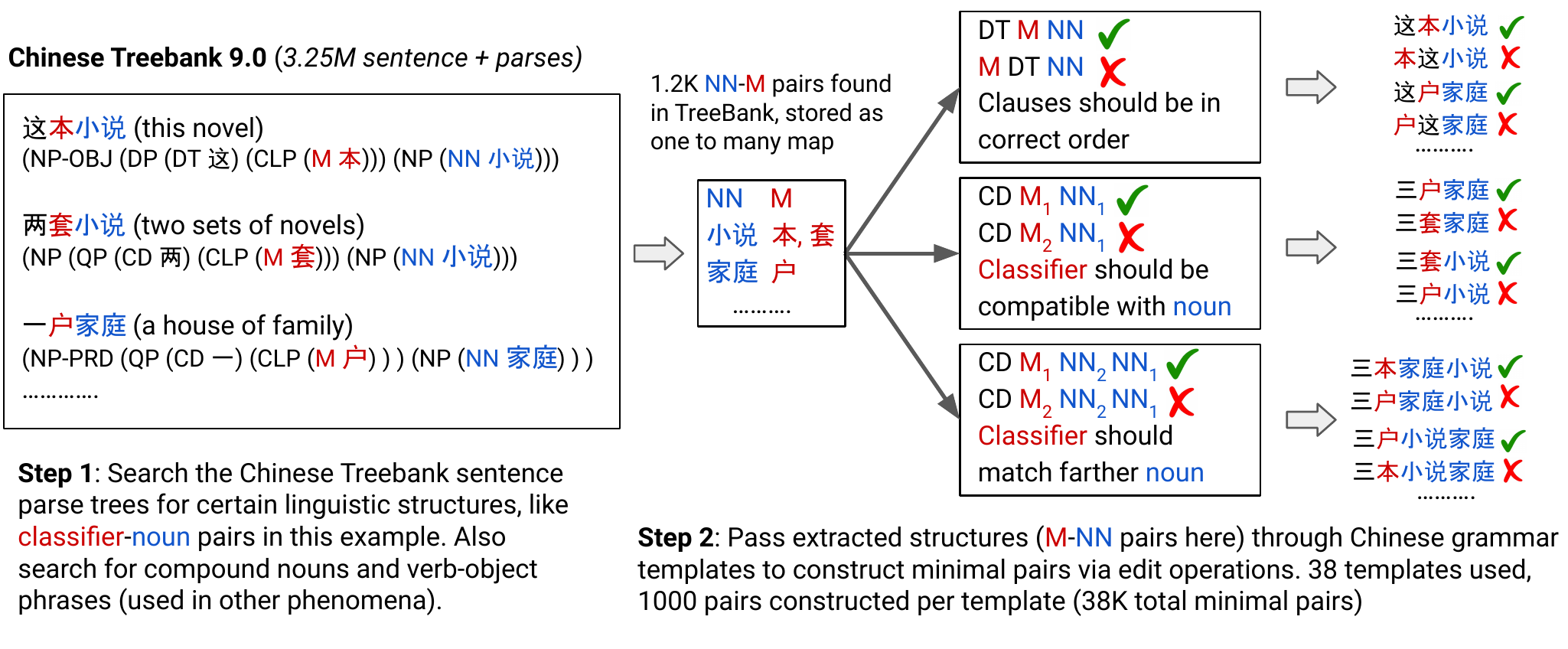}
    \caption{An illustration of the minimal pair generation process used to construct \dataset.}
    \label{fig:sling-generation}
\end{figure*}
  
 \noindent Other issues with these two paradigms are discussed in detail in \appendixref{appen.anaphor}.
  
  \paragraph{Filler-gap paradigm:} To create minimal pairs for the filler-gap paradigm in CLiMP,~\citet{climp} use what they call the topicalization construction. However, (\ref{ex.fillergap}a), taken from CLiMP, does  not contain a filler-gap topicalization dependency. A real topicalization filler-gap structure should be the one in (\ref{ex.fillergap.real}), in which the direct object of the verb \textit{buy} is topicalized and moved to the beginning of the sentence, leaving a {\color{ceruleanblue}\textit{(gap)}} at its base generated position \citep[Section 6.1]{huang2009syntax}. Unfortunately, the minimal pairs associated with this paradigm are generated based on an erroneous template, which means no conclusions can be drawn from model performance on it.
  
    \begin{exe}
    
        \ex\label{ex.fillergap}
        \begin{xlist}
        \ex \begin{CJK*}{UTF8}{gbsn}\small 门，我买了这东西。\end{CJK*}
        
            Door, I bought this thing.
        
        \ex\label{ex.fillergap.real}
            \begin{CJK*}{UTF8}{gbsn}\small 门，我买了 {\color{ceruleanblue}\textit{(gap)}}。\end{CJK*}
        
            Door, I bought {\color{ceruleanblue}\textit{(gap)}}.
            
        \end{xlist}
    \end{exe}

\section{Creating the \dataset~Benchmark }

This section describes our process of generating minimal pairs for \dataset. We make use of the Chinese Treebank 9.0 \citep{ctb9}, a Chinese corpus with linguist-annotated constituency parses that contains 2,084,387 words.
This treebank allows us to use naturally-occurring sentences to construct our minimal pairs, unlike the synthetic and sometimes nonsensical sentences of CLiMP.
Also, unlike CLiMP, whose linguistic templates rely solely on one grammar book \citep{po2015chinese}, our linguistic templates are constructed by a native Chinese linguist (the first author of this paper) based on multiple works in linguistics. Details of the construction of each phenomenon and the cited works can be found in \appendixref{appen.linguistic.phenomena}. 
The general minimal pair generation process is to identify a linguistic pattern, search for relevant linguistic structures in the Treebank, and form minimal pairs by applying hand-crafted transformation rules on the extracted structures. \figureref{fig:sling-generation} provides an overview of this process, with the same running example as this section.

\subsection{Corpus: Chinese Treebank 9.0}

Chinese Treebank 9.0 is a corpus of parsed text (3,247,331 Chinese and foreign characters) from various resources, both formal and colloquial. 
The Treebank contains 132,080 sentences; we extract a subset of these sentences that contains linguistic structures of interest and then manipulate those sentences to create minimal pairs for \dataset. 


\subsection{Pattern Search}\label{sec.pattern.search}

The most important patterns and corresponding strings extracted from the Treebank are classifier-noun phrases, compound noun phrases, and verb-object phrases. To demonstrate the extraction process, we will use classifier-noun phrases as an example. We extract classifier-noun phrases by searching for subtrees that have \texttt{NP} as their root node and contain a classifier \texttt{M}, for example, (\ref{ex.pattern.search}).

\begin{exe}
\small
    \ex\label{ex.pattern.search}
    \begin{CJK*}{UTF8}{gbsn}
    \begin{small}
    \begin{verbatim}
(NP-OBJ (DP (CD 两) 
            (CLP (M 套))) 
        (NP (NN 小说)))\end{verbatim}
    \end{small}
    \end{CJK*}
\end{exe}

\noindent For each sub-tree, a classifier-noun pair is extracted as shown in \figureref{fig:sling-generation}. Because each noun may have multiple compatible classifiers, a dictionary is created with the nouns as keys and the compatible classifiers as the values. Compound noun phrases and verb-object phrases are extracted in a similar way but stored as sub-trees only. 


    

\begin{table*}[!t]
\fontsize{8}{10}\selectfont
\centering 
\resizebox{\textwidth}{!}{%
\begin{tabular}{p{1.5cm}p{6.6cm}p{6.6cm}ccccc}
\toprule
\textbf{Phenomenon}    &  
\textbf{Acceptable Example}  & 
\textbf{Unacceptable Example} & 
\textbf{Syn} & 
\textbf{Sem} & 
\textbf{Distractor} & 
\textbf{Distance} & 
\textbf{Hierarchy}  \\
\midrule
Alternative Question & 
tamen shi laoshi ~~haishi mujiang? \newline they ~~~are teacher or ~~~~~~carpenter \newline ``Are they teachers or carpenters?'' &
tamen shi laoshi ~~haishi mujiang ~~\textbf{ma}? \newline they ~~~are teacher or ~~~~~~carpenter \texttt{SP} &
\ding{51} &           &            &     &     \\

\rowcolor{aliceblue}
 Anaphor \newline (Gender) & 
    nan ~~dianyuan ~~~~~~~kanjianle ~~\textbf{ta}(\begin{CJK*}{UTF8}{gbsn}\tiny 他\end{CJK*})-ziji.\newline male shop assistant saw ~~~~~~~~~~himself\newline
    ``The male shop assistant saw himself."& 
 nan ~~dianyuan ~~~~~~~kanjianle ~~\textbf{ta}(\begin{CJK*}{UTF8}{gbsn}\tiny 她\end{CJK*})-ziji.\newline male shop assistant saw ~~~~~~~~~~herself &
 \ding{51} &           & \ding{51} &  & \ding{51}   \\
 
  Anaphor \newline (Number) & 
    nan ~~dianyuan ~~~~~~~\textbf{men} kanjianle \textbf{tamen}-ziji.\newline male shop assistant \texttt{PL} ~~saw ~~~~~~~~~themselves\newline
    ``The male shop assistants ~saw themselves."& 
 nan ~~dianyuan ~~~~~~~men kanjianle ~~\textbf{ta}-ziji.\newline male shop assistant \texttt{PL} ~~saw ~~~~~~~~~himself &
 \ding{51} &           & \ding{51} &  & \ding{51}   \\
 \rowcolor{aliceblue}
 Aspect      & 
 ta ~\textbf{qu} ~nian zhiding ~~zhengce le.\newline he last year establish policy ~~~\texttt{AS}\newline
 ``He established policies last year."  & 
 ta ~\textbf{ming} nian zhiding ~~zhengce le.\newline he next ~~year establish policy ~~~\texttt{AS} & \ding{51}
  &   &            &     & \ding{51} \\

 Classifier-Noun & 
 yi ~~~\textbf{ming} tielu ~~~~~jingcha\newline one \texttt{M} ~~~~~~~railway policeman\newline 
 ``a railway policeman" & 
 yi ~~\textbf{tiao} tielu ~~~~jingcha\newline one \texttt{M} ~~~railway policeman \newline (\textbf{tiao} is a wrong classifier for \textit{policeman}) & & \ding{51} & \ding{51} & \ding{51} & \ding{51} \\
 \rowcolor{aliceblue}
 Definiteness Effect & 
 zheli/nali ~you ~~\textbf{yi} ~~jia yingyuan.\newline here/there exist one \texttt{M} ~cinema\newline
 ``Here/there exists a cinema." &  
 zheli/nali ~you ~~\textbf{zhe}/\textbf{na}/\textbf{mei} ~jia yingyuan.\newline here/there exist \texttt{DT}/\texttt{DT}/every \texttt{M} ~cinema & 
 &    \ding{51}    &     &    &  \\

 Polarity Item & 
 ta ~~\textbf{bu} ~fazhan ~~renhe youhao ~guanxi.\newline she not develop any ~~~friendly relations \newline
 ``She does not develop any friendly relations."  &
 ta ~~fazhan ~~renhe youhao ~guanxi.\newline she  develop any ~~~friendly relations &
            & \ding{51} &            &       &  \\
\rowcolor{aliceblue}
 Relative Clause  & 
 ta ~~jianle na ge  zhizhile baoli ~~de ~~~n{\"u} ~~~~~~jingcha.\newline
 she saw ~~\texttt{DT} \texttt{M}  ~stoped ~~crime \texttt{DEC} female police\newline
 ``She saw the female police officer who stopped the crime." &
 ta ~~jianle na ge  ta ~~zhizhile baoli ~~de ~~~n{\"u} ~~~~~~jingcha.\newline
 she saw ~~\texttt{DT} \texttt{M} she  ~stoped ~~crime \texttt{DEC} female police & 
 \ding{51} &           &            &  & \ding{51} \\
 \textit{Wh}-fronting & 
 tamen shang ge yue ~~~~daodi ~~~~goujie ~~~~~~~~~~le ~~\textbf{shenme}?\newline they ~~~last ~~~~\texttt{M} ~month on earth collude with \texttt{AS} what\newline
 ``What on earth did they collude with last month?" & 
 \textbf{shenme} tamen shang ge yue ~~~~daodi ~~~~goujie ~~~~~~~~~~le?\newline  what ~~~~they ~~~last ~~~~\texttt{M} ~month on earth collude with \texttt{AS} &
 \ding{51} &           &            &     &   \\
\bottomrule
\end{tabular}}
\caption{An overview of the phenomena present in \dataset\ along with their properties. The table indicates whether the paradigms within each phenomena represent syntactic (syn) or semantic (sem) knowledge, whether they involve a distractor (e.g., the \textit{\textbf{roses}} in the \underline{vase} \textit{\textbf{are}}/*\underline{is} \ldots), whether there are long distance dependencies (e.g., \textit{\textbf{these}} beautiful red blooming \textit{\textbf{roses}}), and whether the LMs need hierarchical knowledge of the language (e.g., \figureref{fig.current.binding.main.text}) to distinguish  acceptable sentences from unacceptable ones. Details of each phenomenon are given in \appendixref{appen.linguistic.phenomena}.}
\label{tab:overview_phenomena}
\end{table*}

\subsection{Sentence Generation}

Minimal pairs are generated based on linguistic templates and the extracted strings. Using the classifier-noun agreement phenomenon as an example, the template is \texttt{CD M Noun}. For the acceptable phrases, the \texttt{M} is taken from the classifiers that are compatible with the noun in the dictionary. For the unacceptable phrases, \texttt{M} is randomly chosen from a classifier list (after making sure it is not in the list of compatible classifiers). 

In addition to phrases extracted from the Treebank, we also extract the transitive verbs\footnote{The transitive verbs from CLiMP are used in a small portion of the minimal pairs in \dataset's \emph{Anaphora} dataset, which requires transitive verbs that take animate subjects and objects. The acceptability contrast of sentences does not rely on those verbs. Extracting such verbs from the Treebank was impossible because animacy of nouns is not encoded in the parse.} used in CLiMP's anaphor and binding phenomena,\footnote{The vocabulary and data generation code of CLiMP can be found here \url{https://github.com/beileixiang/CLiMP}.} and for certain phenomena we also utilize word lists (e.g., locations, pronouns, and occupations) to build the minimal pairs. Finally, for each paradigm in \dataset, we generate one thousand minimal pairs.

\subsection{Phenomena}\label{sec.phenomena}

As summarized in \tableref{tab:overview_phenomena}, \dataset~includes 9 major Chinese linguistic phenomena in syntax and semantics. Several minimal pair paradigms are designed to test an LM's robustness to distance and distractors in a dependency relation as well as whether they have the essential linguistic knowledge of hierarchy in Chinese; more details are provided in \appendixref{appen.linguistic.phenomena}. Here we describe the gist of each phenomenon. The \textbf{alternative question} phenomenon tests the knowledge that the disjunctor \textit{haishi} and the polar question marker \textit{ma} may not co-occur. In the \textbf{anaphor agreement} phenomenon, we first use baselines to test the LMs' gender and number bias (see \appendixref{appen.anaphor}). Then, the morpheme \textit{ziji} (self) is added to test if the LMs knows the function of \textit{ziji} and agree the gender/number of the reflexive with the sentence subject. To avoid the issue caused by Chinese proper names in CLiMP, we use \textit{gender} + \textit{occupation} as the subject of sentences to clearly indicate the gender. The \textbf{aspect} phenomenon tests the knowledge of the perfective aspect markers \textit{le} and \textit{guo} in the sense of their interaction with tense and the progressive marker \textit{zai}. The \textbf{classifier-noun agreement} is observed when a noun is modified by a numeral or demonstrative. One noun can be compatible with more than one classifier and the matching can be idiosyncratic. The \textbf{definiteness effect} phenomenon is established on the observation that demonstrative \textit{zhe} (this)/\textit{na} (that) and the quantifier \textit{mei} (every) may not occur in the post-verbal position of an existential \textit{you} (there is) sentence. \textbf{Polarity item}s (PI) are words or phrases whose occurrence is restricted to certain contexts (e.g., negative or affirmative). We test two negative PIs, \textit{renhe} (any) and \textit{shenme} (what), as well as one positive PI \textit{huoduo huoshao} (more or less). Chinese \textbf{relative clauses} exhibit a filler-gap dependency relationship. If the gap is a simple subject or direct object position, no resumptive noun or pronoun is allowed. Lastly, the \textbf{\textit{wh}-fronting} phenomenon shows that in absence of a specific context (e.g., an echo question), a \textit{wh} phrase must stay in situ. 



\subsection{Human Validation}\label{sec.human.eval}

Two rounds of human validation were conducted on PCIbex \citep{pcibex} to verify the quality of the generated minimal pairs.\footnote{After the first round, the human accuracy on the two compound noun paradigms were $61.36$\% and $77.27$\%. To improve the quality of \dataset, we revised the generation process of the two paradigms and re-evaluated their quality.} Eleven students from the University of Massachusetts Amherst were recruited as annotators for the first round, and five for the second round. Each student has finished at least senior high school in China, and they all use Chinese on a daily basis. For the first round evaluation, every annotator rated 20 pairs from each of the 30 paradigms (not the baselines).\footnote{Ten practice and 24 filler item pairs were created to test whether the annotators understood and paid attention to the task. Those pairs are irrelevant to the paradigms of interest. All annotators did these tests with $100$\% accuracy.} The annotators were shown one minimal pair at a time and asked to choose the more acceptable sentence. In total, the annotation task took 1.5 to 2 hours on average, and the annotators were paid \$$40$ each. Details on the second annotation round can be found in \appendixref{appen.second.human}. The final raw human accuracy mean over all paradigms is $97.12$\% (median = $97.27$\%, SD = $2.29$\%). The inter-annotator agreement as measured by Fleiss' $\kappa$  is $0.8823$, indicating \textit{almost perfect agreement} \citep{landis1977measurement}.


\section{Experimental Setup}

\begin{table}[h]
\centering
\resizebox{\linewidth}{!}{%
\begin{tabular}{llll}
\toprule
LM                           & Param   & Tr. Size & Source  \\
\midrule
\emph{(monolingual models)} \\
{\color{cinnabar} lstm-zh-cluecorpussmall} & 25.8M     & 14G   & \citep{zhao2019uer}\\
{\color{cinnabar} gpt2-zh-cluecorpussmall} & 102M       & 14G      & (same as above) \\
CPM-Generate                 & 2.6B       & 100GB    & \citep{zhang2021cpm}\\
PanGu-$\alpha$               & 2.6B       &  1.1TB   & \citep{zeng2021pangu}\\
bert-base-zh            & 110M       & 25M sent. & \citep{devlin2018bert}\\
{\color{ceruleanblue} zh-pert-base}            & 110M       & 5.4B     & \citep{cui2022pert}\\
{\color{ceruleanblue} zh-pert-large}           & 330M       & 5.4B     & (same as above)\\
{\color{chromeyellow}mengzi-bert-base}             & 103M       & 300G     & \citep{zhang2021mengzi}\\
{\color{chromeyellow}mengzi-bert-base-fin}         & 103M       & 320G     & (same as above) \\
ernie-1.0                    & 110M       & 173M sent.    & \citep{sun2019ernie}\\\midrule\midrule
\emph{(multilingual models)} \\
GPT-3-Davinci      & 175B  &   & \citep{brown2020language}\\
{\color{asparagus}XLM-R-base}             & 270M       &  2.5TB   & \citep{conneau-etal-2020-unsupervised}\\
{\color{asparagus}XLM-R-large}            & 550M       & 2.5TB    & (same as above)\\
BERT-base-multiling-cased & 110M       &          & \citep{devlin2018bert}\\
{\color{bittersweet}MT5-small}                    & 300M       & 26.76TB  & \citep{xue2020mt5}\\
{\color{bittersweet}MT5-large}                    & 1.23B      & 26.76TB  & (same as above)\\
{\color{blue(pigment)}Byt5-small}                   & 300M       & 26.76TB  & \citep{xue2021byt5}\\
{\color{blue(pigment)}Byt5-large}                   & 1.23B      & 26.76TB  & (same as above) \\\bottomrule
\end{tabular}%
}
\caption{The set of Chinese language models evaluated in this work. We consider both large monolingual models and multilingual models (separated by double line). Tr. size = training data size; zh = Chinese; sent. = sentences. Color coded LM pairs were released in the same paper, and differ in size or training data.}
\label{tab:lms}
\end{table}

\noindent \textbf{Evaluated Models: } There are many publicly available pretrained monolingual Chinese LMs and multilingual LMs. While~\citet{climp} only test \model{bert-base-chinese}, three LSTM LMs, and two 5-gram LMs in their work on CLiMP, we experiment with the 18 LMs listed in \tableref{tab:lms}.\footnote{Most LMs tokenize an input sentence into characters but \model{CPM-Generate} and \model{PanGu-$\alpha$} occasionally cuts an input into words, and the \model{ByT5} models use bytes.} There are 6 pairs of LMs (color coded in \tableref{tab:lms}) in which one model is either trained with more parameters than the other in the pair or with larger training data.\footnote{The \model{mengzi-bert-base-fin} model is \model{mengzi-base} further trained with 20G extra financial news and research reports.} Although \model{lstm-zh-cluecorpussmall} and \model{gpt2-zh-cluecorpussmall} also differ in their model structure, we pair them to see whether a Transformer-based architecture leads to  better model performance. We run the same suite of LMs on CLiMP, show the results in \tableref{tab:17CLiMP}, and discuss them in \sectionref{sec.comparison.to.climp}.

\paragraph{Evaluation:} To evaluate the performance of an LM on \dataset, we use perplexity for the causal LMs and pseudo-perplexity \citep{salazar2019} for the masked LMs (see \appendixref{appen.metrics} for details). Given a minimal pair, the LMs should assign a lower (pseudo-)perplexity to the acceptable sentence. The accuracy of each LM on a paradigm is the proportion of the minimal pairs in which the model assigns the acceptable sentence a lower (pseudo-)perplexity. 

\paragraph{Why perplexity?} We choose to use perplexity instead of other metrics (e.g., raw probability) because some phenomena in \dataset~have systematic difference in sentence length within minimal pairs (e.g., \emph{Polarity Item}, \emph{Relative Clause}). Thus, we require a length-normalized metric like perplexity, since metrics such as probability can prefer shorter sentences by nature~\citep{wu2016google, koehn-knowles-2017-six, brown2020language, holtzman-etal-2021-surface}. Additionally, perplexity (or pseudo-perplexity) is applicable to all phenomena and all LMs that are tested in \dataset~(details in \appendixref{appen.metrics}). We considered other evaluation metrics such as prefix methods \citep{2016Linzen,2018Gulordava,2019Wilcox}, by-word surprisal \citep{2018Futrell}, and training an acceptability classifier \citep{2019Warstadt} but eventually decided not to use them for reasons detailed in \appendixref{appen.related.work}.
\section{Results \& Analysis}\label{sec.results}

\begin{table*}[!t]
\centering
\resizebox{\textwidth}{!}{%
\begin{tabular}{@{}l>{\bf}rrrrrrrrrrrrrrrrrrr@{}}

Phenomenon & 
\rotatebox{70}{human} & 
\rotatebox{70}{lstm} & 
\rotatebox{70}{gpt2-zh} & 
\rotatebox{70}{CPM} & 
\rotatebox{70}{PanGu} & 
\rotatebox{70}{bert-base-zh} & 
\rotatebox{70}{pert-base} & 
\rotatebox{70}{pert-large} & 
\rotatebox{70}{mengzi-base} & 
\rotatebox{70}{mengzi-base-fin} & 
\rotatebox{70}{ernie} & 
\rotatebox{70}{xlm-R-base} & 
\rotatebox{70}{xlm-R-large} & 
\rotatebox{70}{bert-base-multi} & 
\rotatebox{70}{mt5-small} & 
\rotatebox{70}{mt5-large} & 
\rotatebox{70}{byt5-small} & 
\rotatebox{70}{byt5-large} & 
\rotatebox{70}{gpt3} \\\toprule
Alternative question & 97.3 & 13.5 & 47.4 & 85.8 & 10.0 & 93.1 & 89.8 & 79.2 & 75.6 & 73.0 & \multicolumn{1}{r|}{94.3} & 53.1 & 56.9 & 6.5 & 45.3 & 10.3 & 25.9 & 55.1 & 14.9 \\
\rowcolor{aliceblue}
Anaphor (gender) & 98.5 & 74.9 & 67.5 & 71.1 & 99.0 & 88.3 & 60.8 & 50.3 & 92.2 & 89.3 & \multicolumn{1}{r|}{81.6} & 59.5 & 61.0 & 82.5 & 50.6 & 37.7 & 53.9 & 37.7 & 63.2 \\
Anaphor (number) & 96.5 & 99.6 & 100 & 92.3 & 0.0 & 99.9 & 99.8 & 98.8 & 80.3 & 75.5 & \multicolumn{1}{r|}{99.5} & 95.2 & 85.2 & 94.7 & 27.3 & 7.3 & 93.6 & 73.0 & 99.9 \\
\rowcolor{aliceblue}
Classifier-noun & 96.4 & 79.9 & 85.7 & 52.7 & 74.8 & 95.3 & 94.9 & 82.2 & 93.9 & 93.5 & \multicolumn{1}{r|}{94.4} & 87.1 & 90.2 & 87.5 & 68.0 & 84.3 & 52.7 & 53.0 & 89.1\\
Aspect & 97.6 & 52.4 & 71.9 & 61.2 & 55.8 & 84.1 & 81.6 & 68.4 & 76.3 & 78.3 & \multicolumn{1}{r|}{74.3} & 54.1 & 68.9 & 45.0 & 49.8 & 65.1 & 55.3 & 50.9 & 71.5 \\
\rowcolor{aliceblue}
Definiteness effect & 96.8 & 97.0 & 99.4 & 70.4 & 68.5 & 96.4 & 95.4 & 73.9 & 96.6 & 96.1 & \multicolumn{1}{r|}{88.7} & 63.5 & 72.8 & 94.1 & 72.2 & 49.0 & 14.2 & 9.0 & 81.5 \\
Polarity item & 92.0 & 90.3 & 86.0 & 78.9 & 79.6 & 72.0 & 90.4 & 94.7 & 97.9 & 98.2 & \multicolumn{1}{r|}{81.3} & 96.5 & 96.5 & 44.2 & 78.2 & 81.6 & 59.5 & 62.9 & 85.9 \\
\rowcolor{aliceblue}
Relative clause & 99.1 & 72.1 & 44.9 & 50.4 & 14.3 & 34.2 & 38.0 & 89.3 & 18.9 & 13.1 & \multicolumn{1}{r|}{33.1} & 43.7 & 48.7 & 13.2 & 42.2 & 50.2 & 2.8 & 18.3 & 65.2 \\
\textit{wh} fronting & 100 & 100 & 99.7 & 93.7 & 94.3 & 99.8 & 99.8 & 99.6 & 99.8 & 99.4 & \multicolumn{1}{r|}{99.8} & 97.4 & 99.4 & 67.8 & 81.1 & 98.6 & 13.1 & 44.7 & 100 \\\midrule
\rowcolor{aliceblue}
Average over phenomena & 97.1 & 75.5 & 78.0 & 72.9 & 55.1 & 84.8 & 83.4 & 81.8 & 81.3 & 79.6 & \multicolumn{1}{r|}{83.0} & 72.2 & 75.4 & 59.5 & 57.2 & 53.8 & 41.2 & 45.0 & 74.6 \\\bottomrule     
\end{tabular}}
\caption{The average percentage accuracy of the LMs and human performance on each phenomenon (random guessing is $50$\%).  Overall, humans significantly outperform all LMs. No LM performs well on all phenomena, but monolingual LMs perform better than multilingual ones. A larger model size does not imply better performance. The vertical line separates the mono/multilingual models. The anaphor phenomenon accuracies include the baselines.}
\label{tab:maintextresult}
\end{table*}

\tableref{tab:maintextresult} reports the human performance and the results of the LMs on each phenomenon.\footnote{The accuracy of each paradigm in all phenomena can be found in \appendixref{appen.results.sling}, along with a visualization in \figureref{fig.all_lm_boxplot}.} Overall, LM performance (\model{bert-base-zh} $84.8$\% being the best) lags far behind human performance ($97.1$\%). Looking into each phenomenon, although some LMs occasionally perform better than humans (e.g., in the definiteness effect), no single LM performs consistently well. Comparing the monolingual LMs to the multilingual ones, the former performs in general better than the latter.\footnote{The poor performance of \model{PanGu-$\alpha$} is partially due to its strong bias toward singular number in the anaphor (number) phenomenon.} In the following subsections, we provide analyses of the model performance from the aspects of model size, distance, and hierarchy. By-phenomenon results and analyses are in \appendixref{appen.by.phenomenon}.

\subsection{Model Size}

To investigate whether a larger model performs better on \dataset, two-tailed pairwise Wilcoxon signed rank tests were conducted on each LM pair in \tableref{tab:lms}. The tests indicated that the performance of the LMs in the \model{pert} and \model{mengzi} LM pairs statistically significantly differed from each other while there is no statistical difference in other LM pairs. Further one-tailed pairwise Wilcoxon signed rank tests on these two pairs revealed (unintuitively) that the smaller LMs (\model{pert-base}, \model{mengzi-base}) perform better than the larger ones (\model{pert-large}, \model{mengzi-fin}). The test results can be found in \tableref{tab:LM.pairs} in \appendixref{appen.analyses}. The finding here coincides with the conclusion drawn in BLiMP and CLiMP that increasing model size does not necessarily improve the model performance.

\subsection{LMs are Affected by Distance}

The classifier-noun phenomenon was designed to test if the LMs are affected by distance in a dependency. For example, in (\ref{ex.cl.noun.distance0}), the {\color{cinnabar}classifier} is separated from the {\color{chromeyellow}noun} by a  {\color{ceruleanblue}long adjective},\footnote{In \dataset, the long adjective is chosen to be eight characters of two conjoined adjectives modified by an adverb \textit{very} as in (\ref{ex.cl.noun.distance0}-\ref{ex.cl.noun.distance1}).} making the local dependency distant. The noun phrase can also be a compound noun (\ref{ex.cl.noun.distance1}), in which case the classifier should agree with the second noun.

\begin{exe}
\small
    \ex\label{ex.cl.noun.distance0}
    \begin{CJK*}{UTF8}{gbsn}三{\color{cinnabar}户}{\color{ceruleanblue}非常优秀且高效的}{\color{chromeyellow}家庭}\end{CJK*}
    
    3 {\color{cinnabar}households} of {\color{ceruleanblue}very excellent and efficient} {\color{chromeyellow}families}
    
    \ex\label{ex.cl.noun.distance1}
    \begin{CJK*}{UTF8}{gbsn}三{\color{cinnabar}本}{\color{ceruleanblue}非常优秀且高效的}{\color{chromeyellow}家庭小说}\end{CJK*}
    
    3 {\color{cinnabar}copies} of {\color{ceruleanblue}very excellent and efficient} {\color{chromeyellow}family fiction}
\end{exe}

Two two-tailed paired Wilcoxon signed rank tests were conducted to compare the simple noun paradigm with and without a long adjective as well as the ones with compound nouns. The results indicated that there was a statistically significant difference between the model performance when the long adjective was present and absent in the simple noun paradigms. There was no such difference in the compound noun paradigm. Further one-tailed Wilcoxon signed rank tests showed that, with a long adjective, the LM performance of the simple noun paradigms decreased. The \textit{p} values are reported in \tableref{tab:discussion.cl.noun.distance}.

\subsection{LMs struggle with Hierarchy}

\begin{figure}[ht]
\centering
\begin{tikzpicture}[scale=0.55]
    \Tree 
    [.S [.DP\textsubscript{1}\\{\color{cinnabar}male scholar} ] 
        [.VP [.PP [.P\\at ] [.DP\textsubscript{2} [.DP\textsubscript{3}\\{\color{ceruleanblue}female director} ] [.D' [.D\\{'s} ] [.NP\\{kiosk} ] ] ] ] 
             [.V' [.V\\{applied-for} ] [.DP\textsubscript{4} [.DP\textsubscript{5}\\{\color{cinnabar}himself\textsubscript{1}} ] [.D' [.D\\{'s} ] [.NP\\{tax return} ] ] ] ] ] ]
\end{tikzpicture}
\caption{The syntax structure of the sentence \begin{CJK*}{UTF8}{gbsn}\small {\color{cinnabar}男学者}在{\color{ceruleanblue}女导演}的店里申请了{\color{cinnabar}他自己}的退税。\end{CJK*} ({\color{cinnabar}The male scholar} applied for {\color{cinnabar}his own} tax return at {\color{ceruleanblue}the female film director}’s shop.) The reflexive anaphor \textit{himself} must be bound by \texttt{DP\textsubscript{1}}, which is hierarchically closer, rather than \texttt{DP\textsubscript{3}}, which is linearly closer. Details of the tree can be found in \appendixref{appen.linguistic.phenomena}. }
\label{fig.current.binding.main.text}
\end{figure}
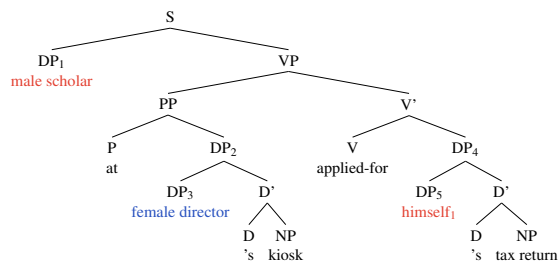

All LMs struggle with hierarchical phenomena and are vulnerable to linear closeness. This is shown in the results for the anaphor and classifier-noun phenomena. The anaphor phenomenon was designed to test whether the LMs prefer linear or hierarchical closeness. For the LMs to correctly choose the acceptable sentences, they should prefer hierarchical closeness. In the example in \figureref{fig.current.binding.main.text}, \texttt{DP\textsubscript{5}} can only agree in its gender feature with \texttt{DP\textsubscript{1}}, which is hierarchically closer. If the LMs are distracted by the linearly closer \texttt{DP\textsubscript{3}}, they would pick the unacceptable sentence in which the \texttt{DP\textsubscript{5}} is \textit{herself}.

Two two-tailed paired Wilcoxon signed rank tests were conducted on the male and female anaphor paradigms with and without a \texttt{PP} respectively. The results show that there is a statistically significant decrease in the performance when the distractor is present.\footnote{This is even the case in the female paradigms where the LMs are strongly biased. The female baseline row in \tableref{tab:18SLNG} shows that when the sentence subject is female, and there is no need for the object to agree with the subject of the sentence, the LMs strongly biased towards a female object. Detailed explanation of the baselines can be found in \appendixref{appen.anaphor}.} The descriptive and test statistics can be found in \tableref{tab:discussion.anaphor.male}.

The classifier-noun phenomenon is designed to test whether the LMs are aware of the right headedness of Chinese compound noun and match the classifier with the second noun in a compound noun rather than the first one (cf. (\ref{ex.cl.noun.distance0}) and (\ref{ex.cl.noun.distance1})). If the LMs do not have this knowledge but prefer linear closeness, they would choose the wrong sentence in a minimal pair. The statistics and the results of two two-tailed Wilcoxon signed rank tests in \tableref{tab:discussion.cl.noun} show that the LMs performed worse when the distractor was present.

\subsection{Strong Gender and Number Bias}\label{sec.gender.number.bias}

Because the LMs can have gender and number bias, in the anaphor phenomena, we use baselines (e.g., \textit{The male baker likes him / her.}) to test the bias.\footnote{The Chinese baseline has the same structure as this English translation.} The higher the accuracy number is, the more biased a LM is towards \textit{him}. \figureref{fig.anaphor.male.baseline} in \appendixref{appen.analyses} shows that, with a male subject, only four monolingual LMs (\model{gpt2-zh}, \model{CPM}, \model{pert-base}, and \model{ernie}) are gender neutral. When the subject is female, all LMs are biased towards a female object (see \figureref{fig.anaphor.female.baseline}). 

One reviewer raised concern that the anaphora resolution in those baselines can only be reliably solved in context of the preceding text, which is true in real life situations. However, in our test setting, since there is no context, the models should ideally be gender neutral on average \citep{bordia-bowman-2019-identifying}.

The LMs also have number bias. A baseline example is \textit{The three male bakers like them / him}. The higher the accuracy number is, the more biased a LM is towards \textit{them}. As seen in the results in \tableref{tab:18SLNG} (\appendixref{appen.analyses}), while most LMs are biased to a plural object when the subject is plural, \model{PanGu-$\alpha$} is strongly biased to a singular object. 

The purpose of the baselines is to reliably test whether the LMs know that the gender/number of \textit{ziji} (self) should agree with the subject's gender/number in the paradigms. As it turn out, the female and number features are not useful for our purpose because the LMs already achieve a `high' accuracy in the baselines, making it ambiguous whether the high accuracy in non-baselines is because they know the function of \textit{ziji} (self) or they are just biased. The male self paradigm, on the other hand, shows that most monolingual LMs were able to use \textit{ziji} as a hint to agree the gender of the subject and object. Among the multilingual LMs, only \model{gpt3-davinci} achieved a meaningful accuracy increase.

\subsection{Vulnerable to Uncertainty}

In the current study, \textit{haishi}, \textit{le}, and \textit{wh} phrases can have more than one usage depending on contexts. The observation is that the LMs performed worse on the paradigms with those phrases. This is most obvious in the aspect and polarity item phenomena. 

In the aspect phenomenon, the possible position of \textit{guo} is relatively fixed compared to \textit{le}, and there is no interaction between \textit{guo} and the progressive marker. The LMs performed better on the \textit{guo} paradigms than on \textit{le}.

In the polarity item phenomenon, the contexts where the positive polarity item \textit{more or less} can occur is more restricted than \textit{any}, which is more restricted to \textit{wh} phrases. And we see that the LM performance is the best on \textit{more or less}, followed by \textit{any}, and the worst on \textit{wh} phrases. 

\subsection{Evaluating Our Set of 18 LMs on CLiMP}\label{sec.comparison.to.climp}

We ran the 18 LMs on CLiMP and compare  model rankings and performance on CLiMP and \dataset. 
We observe major differences: the best LM on \dataset~is \model{bert-base-chinese} ($84.8$\%), and on CLiMP it is \model{chinese-pert-base} ($81.22$\%). That said, monolingual LMs perform better than  multilingual LMs on both datasets.\footnote{Kendall Tau correlation of the two rankings for monolingual LMs is $0.42$ and for multilingual LMs is $0.79$.
} 
While the average performance of the LMs on both datasets is similar (\dataset~$69.7$\%, CLiMP $70.1$\%), on average LMs have significantly larger variation across phenomena on \dataset~(SD = $24.1$\%) than on CLiMP (SD = $13.2$\%). Thus, \dataset~is more discriminative of the strengths and weaknesses of LMs, as LMs tend to be more polarized to one direction across phenomena in \dataset\ compared to those in CLiMP.
Finally, because CLiMP does not test the LMs' bias in the gender and number features for their binding and anaphor paradigms, the LM performance on these two paradigms is uninformative since we do not know what role the bias plays in the tests. \dataset~corrects this issue by including 8 baseline paradigms and shows that the LMs can be strongly biased (see \sectionref{sec.gender.number.bias}). 

\section{Conclusion}

We present \dataset, a new benchmark for evaluating Chinese linguistic knowledge in large scale pretrained LMs. Unlike the existing CLiMP dataset, in which we identify several critical issues, we construct \dataset\ from naturally-occurring sentences in the Chinese Treebank. Our results show that monolingual Chinese LMs achieve better performance on \dataset\ than multilingual LMs. We find that LMs are better at handling local dependencies than long-range dependencies or with distractors, and that they are better at syntactic rather than semantic phenomena. Overall, there remains a large gap between LM and human performance.  

\section*{Limitations}

As a benchmark of evaluating LMs' Chinese linguistic knowledge, \dataset~covers 9 major Chinese grammatical phenomena with 38k minimal pairs. However, there are still phenomena that are important but not included in the current work: for example, the \textit{ba} and \textit{bei} constructions. For those structures, unacceptability can have different sources (e.g., syntax or pragmatics).\footnote{For possible sources of unacceptability of a sentence, please see \citep{abrusan2019semantic}.} Simple syntactic structure restrictions are not enough. When deciding which phenomena to include in \dataset, we deliberately avoid such cases because the (un)acceptability of these phenomena can be mitigated by contextual or world knowledge. As a result, human judgement can vary significantly. As an example, take the \emph{bei} construction (\textit{Passive}): the sentence \begin{CJK*}{UTF8}{gbsn}\small 王萍被嘴举了\end{CJK*} (Wang was lifted by a mouth) is wildly bizarre to some people, while for others, it is acceptable because it is possible to imagine a world in which each body part is a mighty character that can lift things. Such ``unacceptable'' sentences are different from \textit{The roses \textbf{is} red.}, which cannot be resolved by any context. 

Another limitation is that even though Chinese Treebank 9.0 contains a rich and diverse vocabulary, it can still be inadequate at times. For example, for the classifier-noun agreement phenomenon in \dataset, we were not able to extract enough high-quality compound nouns and thus had to manually create 196 minimal pairs, as described in \appendixref{appen.second.human}. One possible way to get around this limitation is to train a parser on the Treebank and use it to automatically parse even more raw Chinese data. We leave this for future work.

\section*{Ethical Considerations}

Following best practices~\citep{mcmillan-major-etal-2021-reusable}, we plan to open source our dataset along with a data card. We will follow the templates used in the GEM benchmark~\citep{gehrmann-etal-2021-gem}\footnote{\url{https://gem-benchmark.com/data_cards}} and HuggingFace Datasets repository~\citep{lhoest-etal-2021-datasets}.\footnote{\url{https://huggingface.co/docs/datasets/v1.12.0/dataset_card.html}} Overall, our project had a small computational cost since we did not need to do any model training. We performed inference on all 18 LMs on a single RTX8000 GPU with 48GB memory. All inference experiments in this paper can be completed within a day on the single GPU.

\section*{Acknowledgements}

First and foremost, we would like to thank all the anonymous reviewers for their valuable comments. We also thank the native Chinese speakers who helped us obtain human performance numbers on \dataset. We are very grateful to Brian Dillon and Simeng Sun for helping formulate the project idea in the early stages of the project. We are also thankful to Yutao Zhou and all the participants in the Semantics Workshop at UMass Linguistics and the UMass NLP group for comments and suggestions during the project. Kalpesh Krishna was supported by the Google PhD Fellowship awarded in 2021.

\bibliography{bib/journal-full,bib/yourbib}
\bibliographystyle{acl_natbib}

\newpage
\appendix

\section{Ngram Count of CLiMP and \dataset}\label{appen.vocab_size}

CLiMP contains 16K minimal pairs (32K sentences) and \dataset~38K (76K sentences). The average sentence length in CLiMP is 11.8 (median = 11) and in \dataset~is 12.5 (median = 12). Because of the difficulty of defining what counts as a word in Chinese, we report one to four ngram counts of types in \tableref{tab:ngram_cnt}, together with the word type counts returned by Jieba.\footnote{\url{https://github.com/fxsjy/jieba}} Because \dataset~has more sentences which can lead to larger type counts, we randomly shuffled the sentences and took 32K sentences to calcuate the ngram and Jieba counts of word types. 

\begin{table}[h!]
\begin{tabular}{@{}llll@{}}
\toprule
      & CLiMP  & \dataset-32K & \dataset-76K \\ \midrule
1gram & 1033   & 2756      & 2886     \\
2gram & 22289  & 33031     & 43122    \\
3gram & 62353  & 64257     & 92972    \\
4gram & 102772 & 87532     & 133900   \\
Jieba & 2335   & 9872      & 11987     \\ \bottomrule
\end{tabular}
\caption{Counts of one to four ngram types in CLiMP and \dataset~and word type counts by Jieba.}
\label{tab:ngram_cnt}
\end{table}

One reason for having 1K sentence pairs in each paradigm is to cancel out the potential influence of word frequency on the perplexity of sentences. Having a diverse vocabulary surely helps in this sense. 

\section{Metrics}\label{appen.metrics}


\paragraph{Causal LMs}

Perplexity (PPL) is used for causal LMs to decide the preferred sentences. Each token $w$ is assigned a probability $p$ given the prefix being seen. The perplexity is calculated based on the log likelihood ($L$). For a sentence of length $m$, its perplexity is calculated as below:
\begin{align*}
    L &= \frac{1}{M}\sum_{i=1}^{m} \log p(w_i | w_{1}...w_{i - 1}) \\
    \text{PPL} &= \text{exp}(-L)
\end{align*}

\noindent Each sentence in a minimal pair is assigned a perplexity value. The one with the lower perplexity is taken as the good sentence that the models choose.

\paragraph{Masked LMs}

Pseudo-perplexity values (pseudo-PPL) are used to evaluate masked LMs \citep{salazar2019}. 
Concretely, tokens in a sentence is masked one after another ($w_j$). The masked language models return a probability distribution over the vocabulary in the masked position given the context surrounding it. For a sentence of length $m$, its pseudo-perplexity is calculated as follows:
\begin{align*}
    w_{\backslash i} &= w_1 ... w_{i-1}, w_{i+1} ... w_{m} \\
    \text{pseudo-}L &= \frac{1}{M}\sum_{i=1}^{m} \log p(w_i | w_{\backslash i}) \\
    \text{pseudo-PPL} &= \text{exp}(-L)
\end{align*}


\section{Related Work: Methods of Evaluating Linguistic Knowledge and Their Limitations in \dataset}\label{appen.related.work}

To investigate what kind of and how much linguistic knowledge large-scale pretrained LMs have compared to human, previous works have focused on limited LMs and probed into the internal encoding of the linguistic knowledge \citep{tenney2019bert, tenney2019you, clark2019does}. Other works investigate the LMs' linguistic knowledge of a small subset of English syntactic grammar by using prefix methods \citep{2016Linzen,2018Gulordava,2019Wilcox}, by-word surprisal \citep{2018Futrell}, or trained an acceptability classifier \citep{2019Warstadt}.

\paragraph{Prefix method} \citet{2016Linzen} focus on English subject-verb dependencies and use a prefix method for evaluation, which requires LMs to assign probabilities to the next word given a prefix. The grammatical next word is expected to have a higher probability (e.g., \textit{The keys \textbf{are}} vs. *\textit{The keys \textbf{is}}). The task includes local subject-verb dependencies (e.g., \textit{The keys \textbf{are}} vs. *\textit{The keys \textbf{is}}) as well as dependencies in distance with distractors (e.g., \textit{The roses in the \underline{vase} by the \underline{door} \textbf{are}} vs. *\textit{The roses in the \underline{vase} by the \underline{door} \textbf{is}}). The prefix method is adopted in later works, for example,  \citet{2018Gulordava} and \citet{2019Wilcox}.

The limitation of the prefix methods is that it mostly applies to inflectional grammatical phenomena in a dependency relationship. For Chinese, a language that largely lacks inflection, the usage of the methods is very limited. Taking \dataset~as an example, the prefix methods are \textit{not} applicable to all nine phenomena because the minimal pairs' acceptability depends on: 

\begin{itemize}
    \item the presence/absence of a crucial word (Alternative Question, Anaphor (number), Aspect, Polarity Item, Relative Clause);
    
    \item the word order (Aspect, \textit{wh} fronting);
    
    \item the choice of a crucial word in the middle of a sentence whose acceptability depends on the part of sentence that is after the word (Anaphor (gender), Classifier-Noun, Definiteness Effect, Polarity Item, Relative Clause).
\end{itemize}

\paragraph{By-word surprisal} Another evaluation method, inspired by the controlled psycholinguistic experimentation, is the by-word surprisal\footnote{Surprisal is the log inverse probability of a word given its prefix \citep{hale2001}.} and sentence completion methods proposed by \citet{2018Futrell} to explore LMs' knowledge of syntax. 
The surprisal reflects whether LMs are affected by the presence/absence of critical words in grammatical configurations. In the sentence completion task, LMs completes a sentence given a prefix. Human annotators then judge the grammaticality of the completed sentences. 

The by-word surprisal method solves one limitation of the prefix methods (i.e., the acceptability depends on the presence/absence of a crucial word) but still does not account for the other two listed above. The sentence completion method faces similar restrictions and cannot be applied in a large scale because it requires human judgement of the completed sentences.

\paragraph{Acceptability classifier} \citet{2019Warstadt} trained an acceptability classifier to perform a grammaticality judgement task, which consists of sentences collected from the linguistics literature marked for their acceptability. 

There are several limitations of training a classifier. First, it involves many debatable design decisions (e.g., hyper-parameters). Second, LMs may learn the task from the training data \citep{hewitt-liang-2019-designing, voita-titov-2020-information}. Our goal is to measure the linguistic capability of \textit{pretrained} LMs without additional help from a training dataset that has the same distribution as the test set.

Overall, the previous methods are either only applicable to a subset of linguistic grammar or depend on the performance of a classifier. The minimal pair method used in BLiMP breaks through these limitations. 

\paragraph{Minimal pair method} To cover a wide range of linguistic phenomenon, \citet{blimp} introduced minimal pair evaluation for LMs and created the Benchmark of Linguistic Minimal Pairs for English (BLiMP). It evaluates the linguistic knowledge of twelve English grammatical phenomena including syntax and semantics. Each of them consists of minimal pair paradigms representing different aspects of the phenomena. All minimal pairs are code-generated using templates created by linguists and an annotated vocabulary that contains 3000 words. The dataset is human validated. 

The results on BLiMP show that the LMs tested in BLiMP are good at local dependency relations (e.g., morphology agreement) but bad at phenomena involving hierarchy and semantic knowledge. 
Concerning the training size and model size, while increasing training size can improve model performance, increasing model size does less so. 

\paragraph{Other possible metrics and their limitations} Other possible metrics are probability and a masked-token method. However, probability is not a suitable metric to use in \dataset~for at least two reasons. First, probability is only useful for minimal pairs whose sentences have the same length. Otherwise, probability by nature prefers shorter sentences. Second, the sentences in a minimal pair need to have similar word orders. This is because tokenizers might tokenize a sentence in different ways depending on the word order, causing the sentence length of the sentences in a minimal pair to be different. In the masked-token method, we can mask out the crucial word in each sentence in a minimal pair and ask a LM to give probability of the two masked words. This method is not applicable to causal LMs. For masked LMs, it is only applicable to Anaphor (gender), Classifier-Noun, and Definiteness Effect in \dataset~where the word order does not change. In those cases, since \dataset~uses minimal pairs, the masked token in those phenomena will be exactly the part in which the sentences in a minimal pair differ. Hence, the masked-token method will return the same results as the pseudo-perplexity.

\section{Linguistic Phenomena}\label{appen.linguistic.phenomena}

The current work focuses on six syntax and three semantics phenomena in Chinese. \tableref{tab:overview_phenomena} offers an overview. There are 30 test paradigms. The anaphor phenomenon has 8 baseline paradigms to detect LMs' gender (male/female) and number (singular/plural) biases. 

 All phenomena have at least one paradigm that can be solved by checking the linear order of tokens. Some phenomena require a negative co-occurrence of words. For example, in the alternative question phenomenon, the disjuntor \textit{haishi} and the polar question particle \textit{ma} may not co-occur. Other phenomena require a positive co-occurrence. For example, in the polarity item phenomenon, the grammaticality of \textit{renhe} (any) dependes on the occurrence of negation.
 
 Three phenomena contain paradigms that require the LMs to use the knowledge of hierarchy. If LMs use linear closeness rather than hierarchical closeness, they will wrongly assign a lower perplexity to the unacceptable sentence in a minimal pair. The anaphor phenomenon, for example, contains such paradigms. 
 
 The anaphor, classifier-noun agreement, and relative clause phenomena have paradigms that test LMs' robustness to distractors and long distance dependencies. A distractor is an element that intervenes between the head and its dependent in a dependency/agreement relation. For example, in \textit{The \textbf{roses} in the \underline{vase} \textbf{are}} \ldots, \textbf{roses} and \textbf{are} are in a dependency relation, and \underline{vase} is the distractor. By distance, it is meant to be the case that the head and its dependent is separated from each other (e.g., \textit{\textbf{these} beautiful red blooming \textbf{roses}}).

This section introduces phenomena in turn. If a phenomenon is in CLiMP, a comparison between CLiMP and the current work will be provided.

\subsection{Alternative Questions with \textit{haishi}}\label{appen.haishi}

Chinese alternative questions (AltQ) are most reliably marked by the disjunctor \textit{haishi} \citep{huang2009syntax}. Although \textit{haishi} has different usages \citep{haishi}, when it is used as the disjunctor, the polar question particle \textit{ma} (\texttt{SP}) cannot occur. Minimal pairs like (\ref{ex.altq}) test whether LMs are aware of this. The paradigm concerns only linear co-occurrence.\footnote{The notation \textit{(*ma)} in (\ref{ex.altq}) means that the sentence is good without \textit{ma} but bad with it.}

        

\begin{exe}
    \ex\label{ex.altq}
    \gll tamen shi laoshi haishi mujiang (*ma)?\\
        they are teacher or carpenter (*\texttt{SP})\\
        \glt ``Are they teachers or carpenters?"
\end{exe}

\subsection{Anaphor}\label{appen.anaphor}

Mandarin Chinese has two reflexive pronouns: \textit{ziji} and \textit{ta(men)-ziji}. The former is morphologically simple with no person, number, or gender features. The latter, \textit{ta(men)-ziji}, has the pronoun \textit{ta} which encodes gender features in writing: \begin{CJK}{UTF8}{gbsn}\small 她\end{CJK} for singular female third person, \begin{CJK}{UTF8}{gbsn}\small
他\end{CJK} for singular male third person, and \begin{CJK}{UTF8}{gbsn}\small 它\end{CJK} for singular non-human third person. The character \textit{men} indicates plurality. Because of this morphological richness, \textit{ta(men)-ziji} is used to form minimal pairs. Since CLiMP contains the binding phenomenon, their implementation will be first introduced, followed by the binding phenomenon in the current work.

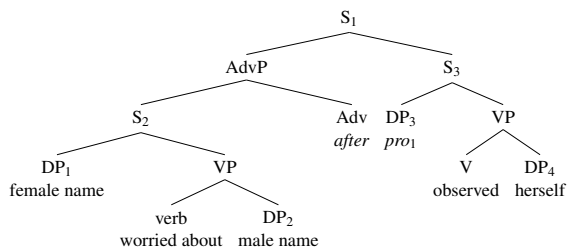
\begin{figure}[ht]
\centering
\begin{tikzpicture}[scale=0.63]
    \Tree 
    [.S\textsubscript{1} [.AdvP [.S\textsubscript{2} [.DP\textsubscript{1}\\{female name} ] 
                                    [.VP [.verb\\{worried about} ] [.DP\textsubscript{2}\\{male name} ] ] ] 
                                [.Adv\\\textit{after} ] ] 
                         [.S\textsubscript{3} [.DP\textsubscript{3}\\\textit{pro}\textsubscript{1} ] 
                                              [.VP [.V\\{observed} ] [.DP\textsubscript{4}\\herself ] ] ] ]
\end{tikzpicture}
\caption{The syntax structure of (\ref{ex.climp.binding}).}
\label{fig.climp.binding}
\end{figure}

\paragraph{Binding Phenomenon in CLiMP}\label{sec.binding.climp} \citet{climp} use singular female and male third person reflexives \textit{ta-ziji} to test the LMs' knowledge of binding. There are two paradigms. The first one has a simple SVO structure in which the object is an anaphor and needs to match the gender feature of the subject. 
The second paradigm involves a distractor between the antecedent and the reflexive (e.g., \texttt{DP\textsubscript{2}} in \figureref{fig.climp.binding}). The distractor is different from the true antecedent in its gender feature. The distractor is linearly closer to the reflexive but hierarchically farther. It turns out that the LMs struggle with this paradigm. The results show that the LMs did no better than chance. One of the acceptable binding sentences in \citet{climp} is cited below. We provide its syntax in \figureref{fig.climp.binding}. The corresponding unacceptable sentence changes \textit{herself} to \textit{himself}. 

\begin{exe}
    \ex\label{ex.climp.binding} 
    \gll {Huang Xiuying} danxin {Wang Hao} zhihou guanchaguo ta-ziji.\\
    female.name worry-about male.name after observe herself\\
    ``After Huang Xiuying worried about Wang Hao, she observed herself."
\end{exe}

\vspace{-0.5\baselineskip}
\vspace{-0.3\baselineskip}

Although, by comparing the two paradigms, \citet{climp} find the models are bad at dealing with hierarchy and distractors, there are four shortcomings in the minimal pair design that weaken the strength of the observation. First, it was not tested whether the LMs knew the gender of the proper names. Because Chinese names do not always clearly indicate the gender, this can cause the LMs guessing randomly. Second, the syntax of the second paradigm is complex because it involves ellipsis.\footnote{The ellipsis is presented as \textit{pro}\textsubscript{1} in \texttt{DP\textsubscript{3}} in Figure \ref{fig.climp.binding}. The index $1$ indicates its antecedent is \texttt{DP\textsubscript{1}}.} With the presence of ellipsis, it is not for sure that the models did bad because they preferred a linearly closer agreement or because they couldn't recover the omitted subject correctly. Third, CLiMP does not have a baseline for the gender biases of the LMs. Hence, we cannot know if the models know the function of \textit{ziji} or they simply prefer one gender. Fourth, CLiMP does not have separate corpora for the two genders. Thus, we do not know if the LMs are bad in both female and male reflexive agreements or only in one of them. 


\paragraph{Paradigms in Current Work}

To amend the four shortcomings, the current work includes baseline paradigms to test LMs' gender bias. Sentences have a simple SVO structure. Instead of using proper names as the subject, the paradigms use gender plus occupations to indicate the gender of a noun. The female and male reflexive agreements are tested separately. 

To form the baseline minimal pairs for the male reflexive agreement, an occupation and a transitive verb were chosen randomly. Following the verb is either a male or female pronoun. Example (\ref{ex.current.binding}) is one resulting minimal pair.

\begin{exe}
    \ex\label{ex.current.binding}
       \gll nan dianyuan baituole \textbf{ta} / \textbf{ta}.\\
                male {shop assistant} {got rid of} him / her\\
            \glt ``The male shop assistant got rid of him."
\end{exe}

\noindent Both sentences are acceptable. The purpose is to see whether the models are gender biased when there is no clue for any gender agreement. Other baselines are formed in the same way.

With the baseline being established, the minimal pairs for the reflexive agreement are created by adding \textit{ziji} to the end of the sentences in the baselines. This turns (\ref{ex.current.binding}) into (\ref{ex.current.binding.reflexive}). Because the presence of \textit{ziji}, the gender of \textit{ta} should agree with the gender of \textit{the male shop assistant}. Hence, \textit{himself} is acceptable but \textit{herself} is not. Such agreement can be solved by linear closeness.

            

\begin{exe}
    \ex\label{ex.current.binding.reflexive}
    \gll nan dianyuan baituole \textbf{ta}- / *\textbf{ta}-ziji.\\
                male {shop assistant} {got rid of} him- / *herself\\
    \glt {\small ``The male shop assistant got rid of himself."}
\end{exe}

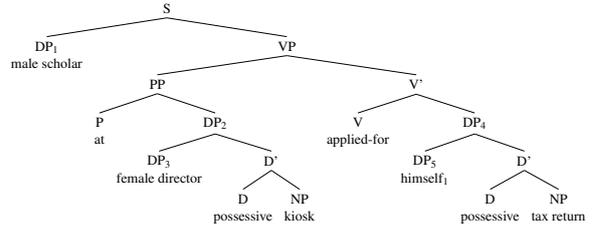
\begin{figure}[ht]
\centering
\begin{tikzpicture}[scale=0.48]
    \Tree 
    [.S [.DP\textsubscript{1}\\{male scholar} ] 
        [.VP [.PP [.P\\at ] [.DP\textsubscript{2} [.DP\textsubscript{3}\\{female director} ] [.D' [.D\\possessive ] [.NP\\{kiosk} ] ] ] ] 
             [.V' [.V\\{applied-for} ] [.DP\textsubscript{4} [.DP\textsubscript{5}\\himself\textsubscript{1} ] [.D' [.D\\possessive ] [.NP\\{tax return} ] ] ] ] ] ]
\end{tikzpicture}
\caption{The syntax structure of the sentence in (\ref{ex.current.binding.hierarchy}) with \textit{himself} being bound by \texttt{DP\textsubscript{1}}.}
\label{fig.current.binding}
\end{figure}

The next paradigm tests whether LMs prefer a linearly closer or a hierarchically closer noun as the antecedent of an anaphor. An example is (\ref{ex.current.binding.hierarchy}). The syntax of the grammatical sentence is in Figure \ref{fig.current.binding}.

\begin{exe}
    \ex\label{ex.current.binding.hierarchy}
       \gll nan xuezhe zai n\"u daoyan de dian shenqingle \textbf{ta}- / *\textbf{ta}-ziji de tuishui.\\
                male {scholar} at female director \texttt{DEG} shop applied-for him- / *herself \texttt{DEG} {tax return}\\
            \glt ``The male scholar applied for his own tax return at the female film director's shop."
            
\end{exe}


Like Figure \ref{fig.climp.binding}, Figure \ref{fig.current.binding} involves a distractor \texttt{DP\textsubscript{3}} but has no ellipsis. It is a SVO sentence with a preposition phrase (\texttt{PP}) modifying the verb phrase. The antecedent of \texttt{DP\textsubscript{5}} can only be \texttt{DP\textsubscript{1}} which c-commands \textit{himself} while \texttt{DP\textsubscript{3}} is embedded deeply in \texttt{PP}. \texttt{DP\textsubscript{1}} is hierarchically closer to \textit{himself} while \texttt{DP\textsubscript{3}} is linearly closer. The LMs will fail if they have no knowledge of hierarchical structure. 

The current work also uses the number feature to test LMs. Baselines are used to see if the tested LMs are biased to singularity or plurality. The gender feature is kept constant so that any distinct behaviour is only caused by the number feature.

\subsection{Aspect Marker \textit{le} and \textit{guo}}\label{appen.aspect}

The morphemes \textit{le} and \textit{guo} often function as perfective aspect markers.\footnote{For the other usages of \textit{le} and \textit{guo}, see \citet{huang2009syntax}, \citet{wang2002motion}, and \citet{pan2004role}, among others.} Although they can occur in sentences of various tenses, without the help of a future oriented adverb together with morphemes as \textit{cai} or \textit{jiu}, they only occur in sentences of past tenses. A paradigm is built on this observation. An example is in (\ref{ex.aspect.temporal}).

\begin{exe}
    \ex\label{ex.aspect.temporal}
    \gll ta \textbf{qu} / *\textbf{ming} nian zhiding zhengce le.\\
    he last / *next year establish policy \texttt{AS}\\
    \glt ``He established policies last year."

\end{exe}

The next paradigm is based on a restriction on \textit{guo} that it cannot co-occur with the progressive marker \textit{zai}, as in (\ref{ex.aspect.prog}).

\begin{exe}
    \ex\label{ex.aspect.prog}
        \gll tamen zai shi (*\textbf{guo}) na ge fuwu.\\
        they \texttt{AD} try (*\texttt{AS}) \texttt{DT} \texttt{M} service\\
        ``They are trying out that service."
        
\end{exe}

The above paradigms can be solved linearly but the interaction between \textit{le} and \textit{zai} requires the knowledge of hierarchy. The morpheme \textit{le} can co-occur with \textit{zai} if \textit{le} takes scope over \textit{zai} but not the other way. Based on this, two paradigms are formed. The first one (\ref{ex.le.prog.le}) tests the knowledge that \textit{le} cannot scope under \textit{zai}. The other paradigm (\ref{ex.le.wide}) shows that \textit{le} can scope over \textit{zai}.

\begin{exe}
    \ex\label{ex.le.prog.le}
        \gll tamen zai guancha (*\textbf{le}) xuanju. \\
        they \texttt{AD} observe (*\texttt{AS}) election\\
        ``They are observing the election."
        
    
    \ex\label{ex.le.wide}
    \begin{xlist}
        \ex \gll tamen zai jiao fakuan \textbf{le}. \\
        they \texttt{AD} pay fine \texttt{AS}\\
        ``They are (already in the process of) paying the fine."
        
        \ex[*]{\gll tamen zai jiao \textbf{le} fakuan. \\
        they \texttt{AD} pay \texttt{AS} fine \\}
    \end{xlist}
\end{exe}

\subsection{Classifier-Noun Agreement}\label{appen.cl-noun}

Classifiers are pervasive in Mandarin Chinese.\footnote{In the current paper, the word `classifier' is used as a cover term for both classifiers and measure words. For the differences between classifiers and measure words, interested readers can refer to \citet{wang1994cls}.} They match with nouns and indicate in what unit a noun is quantified \citep{huang2009syntax}. The difficulty in classifier-noun agreement is that the matching can be idiosyncratic, and one noun can be compatible with multiple classifiers. 

CLiMP includes the classifier-noun agreement phenomenon which consists of three paradigms. However, because the variables in their minimal pairs are not well controlled, the experiment results are not conclusive.

\paragraph{Classifier-Noun Agreement in CLiMP}\label{sec.cl.noun.climp}

Their first paradigm is the local classifier-noun matching. The second paradigm inserts an adjective with two to four characters between the classifier and the noun to increase the distance of the two. There is no distractor in the adjective. The third paradigm further increase the distance by having a relative clause instead of an adjective. Without showing the results of each paradigm, \citet{climp} 
report that the mean of the model performance is 71.66\% (median 70.1\%). Chinese BERT performs the best (92.9\%). The overall human accuracy of the paradigms is 99.7\%.

There are two issues with the paradigms. First, some minimal pairs do not show a clear contrast. Example (\ref{ex.climp.cl}) is taken from CLiMP, in which the classifier \textit{jia} is intended to be unacceptable. However, both \textit{liang} and \textit{jia} are compatible with the noun \textit{bike}.

\begin{exe}
    \ex\label{ex.climp.cl}
        \gll {Sun Yingying} zhengzai reng yi \textbf{liang} / *\textbf{jia} zixingche.\\
        {female name} \textsc{Prog} throw one \texttt{M} / *\texttt{M} bike\\
        \glt ``Sun Yingying is throwing a bike."
        
\end{exe}

\noindent The reason for the issue is that each noun in the CLiMP vocabulary is associated with only one classifier. However, as mentioned before, the classifier-noun matching can be a many to many relation. The second issue is the relative clauses in the third paradigm. Some relative clauses contain a distractor. In certain cases, the distractor even matches the classifier. 

\paragraph{Paradigms in Current Work}

The current work has five paradigms for the classifier-noun agreement. To avoid the issues in CLiMP, we built a classifier-noun dictionary. Each noun is associated with a group of classifiers. When creating the minimal pairs, it is ensured that the classifier in the unacceptable sentences is not listed as a compatible classifier of the noun. 

In the five paradigms, one paradigm tests models' knowledge of the linear order of demonstratives (\texttt{DT}) or numerals (\texttt{CD}) and classifiers (\texttt{M}) before a noun. The other four paradigms test LMs' knowledge of classifier-noun agreement. 

The first of the four paradigms involves local classifier-noun agreement. The second paradigm inserts a long adjective between the classifier and the noun but, still, no knowledge of hierarchy is needed. The third paradigm is based on compound nouns. An example is given in (\ref{ex.compnoun}). 

\begin{exe}
    \ex\label{ex.compnoun}
    
      \gll yi \textbf{ming} / *\textbf{tiao} tielu jingcha\\
            one \texttt{M} / *\texttt{M} railway policeman\\
        \glt ``a railway policeman"
        
\end{exe}

\noindent A Chinese compound noun can be formed by two nouns, \texttt{noun1} (\textit{railway}) and \texttt{noun2} (\textit{policeman}), with \texttt{noun1} modifying \texttt{noun2}. The classifier agrees with \texttt{noun2} \citep{huang2009syntax}. Hence, \texttt{noun1} functions as a distractor. In (\ref{ex.compnoun}), \textit{ming} is the classifier for \textit{policeman} while \textit{tiao} is for \textit{railway}. The last paradigm adds a long adjective after the classifier in the third paradigm. For the compound noun paradigms, the knowledge of hierarchy is needed. That is, the LMs should know the right-headedness of Chinese compound nouns.

\subsection{Definiteness Effect}\label{appen.def.effect}

It has long been noticed that certain strong determiners cannot be in the postverbal position in an English existential \textit{there}-sentence \citep{keenan1987definiteness,abbott1993pragmatic,zucchi1995ingredients}. Similar effects have been observed in Chinese \citep{xu1995definiteness,hu2008focus}. The phenomenon to be tested here involves Chinese \textit{you} (have), a close counterpart to the \textit{there}-construction. The demonstratives \textit{zhe} (this) and \textit{na} (that) as well as the quantifier \textit{mei} (every) are used as an equivalence to the strong determiners in English. The phrase \textit{yi} (one) + \texttt{M} is used as a counterpart of English weak determiners. This paradigm can be solved by checking the linear co-occurrence of two elements, \textit{here}/\textit{there} and the strong determiners. An example is in (\ref{ex.def.effect1}).

\begin{exe}
    
    \ex\label{ex.def.effect1}
    \begin{xlist}
        \ex 
        \gll zheli/nali you \textbf{yi} jia yingyuan.\\
        here/there exist one \texttt{M} cinema\\
        ``Here/there exists a cinema."
        {\small
        \ex[*]{\gll zheli/nali you \textbf{zhe}/\textbf{na}/\textbf{mei} jia yingyuan.\\
        here/there exist \texttt{DT}/\texttt{DT}/every \texttt{M} cinema\\}}
    \end{xlist}
\end{exe}

\subsection{Polarity Items}\label{appen.PI}

Polarity items (PI) are common in natural languages \citep[a.o.]{toth1999negative,yoshimura2007focus,kumar2013syntax,giannakidou2019negative}. English, for example, has \textit{any}, \textit{ever}, and \textit{yet}, etc. In Chinese, \textit{renhe} (any) and \textit{shenme} (what) are two actively investigated negative PIs. They occur in negation, polar questions, and conditionals \citep{cheng1994wh,wang1996syntactic,lin1998existential,chen2012chinese,lin2015no}. The phenomenon contains three paradigms. There is no complex hierarchical structure involved. 
All paradigms can be solved by just checking the linear co-occurrence or absence of certain tokens. The first one concerns \textit{renhe} (any). The acceptability contrast is established by the presence of negation.\footnote{The notation \textit{*(\textbf{bu})} means that the sentence is unacceptable without \textit{bu}.} 

\begin{exe}
    \ex\label{ex.polarity.any}
            \gll ta *(\textbf{bu}) fazhan renhe youhao guanxi.\\
                she not develop any friendly relations \\
            \glt``She does not develop any friendly relations."
            
\end{exe}

The second paradigm involves \textit{shenme}, a multi-functional phrase. It is often seen in \textit{wh}-questions (e.g., \textit{ni}\textsubscript{you} \textit{chi}\textsubscript{eat} \textit{shenme}\textsubscript{what} ``what do you eat?"). However, \textit{shenme} also occurs in the contexts where typical negative PIs occur. The acceptability contrast is manipulated by the presence of negation. Yet, to avoid a \textit{wh}-question reading, the adverb \textit{shenzhi} (even) is used, which can occur in affirmative or negative contexts but not in \textit{wh}-questions as it can be a focus intervener \citep{beck2006intervention}.

\begin{exe}
    \small
    \ex\label{ex.polarity.shenme}
            \gll tamen shenzhi *(\textbf{mei}) sheji shenme liyi.\\
                they even not involve what interests \\
            \glt``They weren't even involved in any interests."
            
\end{exe}

The last paradigm in the current phenomenon focuses on the adverb \textit{huoduo huoshao} (more or less). It is less studied than \textit{renhe} (any) or \textit{shenme} (what). Nonetheless, by searching in the corpus CCL\footnote{CCL is a Chinese corpus curated by Center for Chinese Linguistics at Peking University. It contains 581,794,456 characters in its Contemporary Chinese corpus. Text sources include transcribed spoken language, newspaper, practical writing, literature, etc. Details can be found at \url{http://ccl.pku.edu.cn:8080/ccl_corpus/corpus_statistics.html}.}, it is confirmed that there is no sentence in which \textit{bu} or \textit{mei} (not) negates the verb within 10 characters before or after \textit{huoduo huoshao}. Hence, the acceptability of the minimal pairs is built on the absence of negation.\footnote{The minimal pairs of this paradigm differ in two aspects. First, the acceptable sentences contain \textit{le} but the unacceptable ones do not. Second, the acceptable sentences do not contain \textit{mei} but the unacceptable ones do. This seems render the pairs not minimally distinct. However, the morpheme \textit{mei} is a negation that encodes the perfective aspect. This is what \textit{le} does in the acceptable sentences. Keeping \textit{le} in the unacceptable sentences will make them unacceptable for a reason that is not at issue here. Hence, even though on the surface the two sentences are not minimally distinct, they semantically are.}

\begin{exe}
    \small
    \ex\label{ex.polarity.moreorless}
            \gll tamen huoduohuoshao (*\textbf{mei}) fadong le jingong.\\
                they more-or-less (*not) start \texttt{AS} attack \\
            \glt``They more or less started the attack."
            
\end{exe}

\subsection{Relative Clauses}\label{appen.rc}

Relative clauses in Mandarin Chinese are head-final, meaning a modifying clause occurs before a modified noun. This characteristic is tested in CLiMP. Another characteristic of Chinese relative clauses is that it is a filler-gap construction and, in the gap position, a resumptive noun is out of the question, and a resumptive pronoun cannot occur freely. As cited in \citet{wen2020relative}, \citet{zhou2012phase} point out that resumptive pronouns may not occur in simple subject or direct object positions. The current study 
uses this property and constructs minimal pairs as in (\ref{ex.rc.noun}). If the LMs are not aware of the relative clause structure in those sentences, they can perform poorly because of the local coherence created by the filled-in gaps. 

\begin{exe}
    \small
    \ex\label{ex.rc.noun}
            \gll ta jiandao le na ge (*\textbf{n{\"u}} \textbf{jingcha} / \textbf{ta}) zhizhi le baoli de n{\"u} jingcha.\\
                she see \texttt{AS} \texttt{DT} \texttt{M} (*female police / she) stop \texttt{AS} violence \texttt{DEC} female police\\
            \glt``She saw the female police officer who stopped the violence."
            
\end{exe}

            

\subsection{\textit{Wh}-fronting}\label{appen.wh}

As mentioned in \sectionref{appen.PI}, \textit{shenme} is frequently used to form \textit{wh}-questions. In canonical \textit{wh}-questions, the \textit{wh}-phrases stay in situ \citep{huang2009syntax}. Without a very specific appropriate context, \textit{wh}-fronting is unacceptable. Hence, no matter whether \textit{shenme} alone functions as an object or modifies a noun as in (\ref{ex.wh.bare}), the noun phrase containing it cannot be fronted. To force a question reading of \textit{shenme}, the phrase \textit{jiujing} or \textit{daodi} (on earth) are added. There is no complex hierarchy in the sentences and the \textit{wh} phrases are all objects. 

\begin{exe}
    \small
    \ex\label{ex.wh.bare}
    \begin{xlist}
        \ex \gll tamen shang ge yue daodi goujie le shenme (heidao)?\\
                they last \texttt{M} month {on earth} {collude with} \texttt{AS} what mobster\\
            \glt``What (mobster) on earth did they collude with last month?"
            
        \ex[*]{\gll shenme heidao tamen shang ge yue daodi goujie le?\\
                what mobster they last \texttt{M} month {on earth} {collude with} \texttt{AS}\\}
    \end{xlist}
\end{exe}

            

\section{Second Round of Human Validation}\label{appen.second.human}

The minimal pairs of the two compound noun paradigms were refined. Among the 2000 new minimal pairs, 1804 were code generated and 196 were manually created. To verify the minimal pair quality, a second round of human validation was conducted. Five annotators (3 female, 2 male) with an average age of $22.2$ were recruited the same way as described in \sectionref{sec.human.eval}. 

Twenty pairs of sentences were randomly sampled from both the code generated and manually created minimal pairs from each paradigm. The practice and filler items were used. Each annotator rated 114 pairs. They did the practice and filler items with $100$\% accuracy. The task took less than $10$ minutes. The annotators were paid \$$5$. The raw accuracy on the new validated pairs was $95.25$\% ($\kappa$ = $0.8823$). The manually created minimal pairs had a higher accuracy than the code generated ones ($97.5$\% \textit{vs}.\ $93$\%). After the second round, the raw human accuracy mean over all paradigms is $97.12$\%.

\section{By-phenomenon Results and Analyses}\label{appen.by.phenomenon}

\noindent\textbf{AltQ} The multi-lingual LMs either prefer the sentences with \textit{ma} or perform near chance. Although the mono-lingual LMs perform better, only \model{bert-base-zh} and \model{ernie} have an accuracy higher than $90$\%. There can be multiple reasons for the unsatisfactory performance. First, \textit{haishi} is multi-functional, which might cause the LMs being unsure of its disjunctor usage. Second, \textit{ma} only occurs in interrogative contexts, which can make the LMs prefer having it. Third, the LMs do not have a global 
view of the sentences but only attend to parts of them, which can be the reason of their random guessing.\footnote{The \textit{A haishi B} disjunction and \textit{ma} being at the end of a question are both locally grammatical. } 

\begin{figure*}[!h]
\centering
\resizebox{\textwidth}{!}{%
\includegraphics{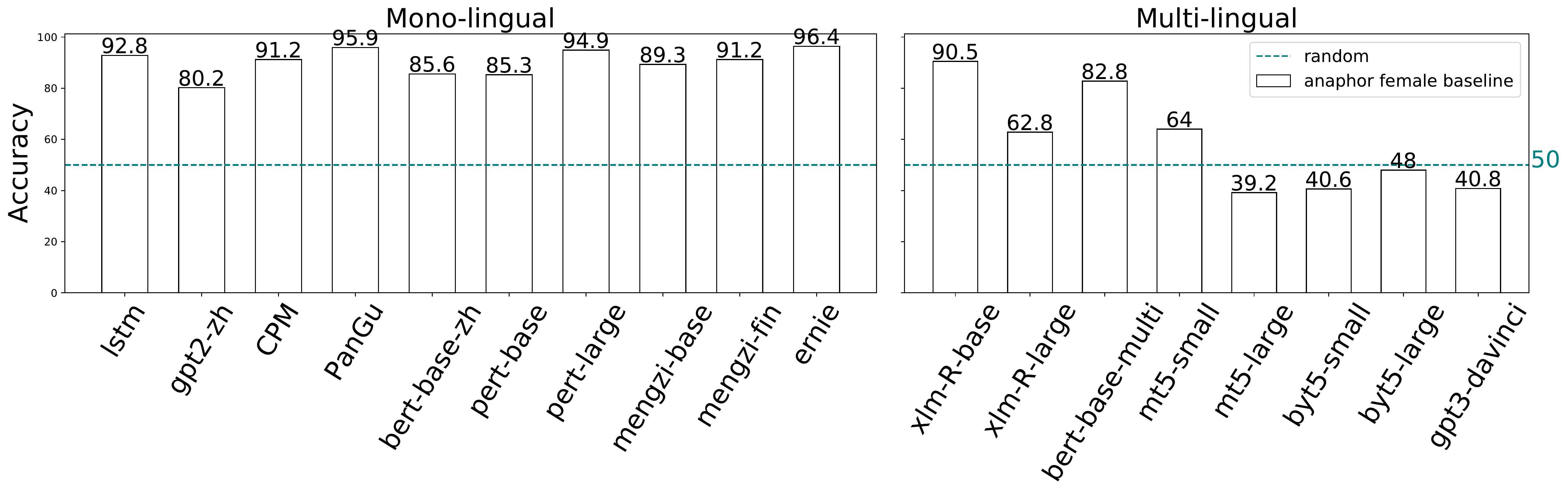}%
}
\caption{The LM bias towards a female object when the subject is female.}
\label{fig.anaphor.female.baseline}
\end{figure*}

\noindent\textbf{Anaphor (Gender)} The LMs are gender biased. \figureref{fig.anaphor.male.baseline} shows that, with a male subject, only four mono-lingual LMs (\model{gpt2-zh}, \model{CPM}, \model{pert-base}, and \model{ernie}) are gender neutral. When the subject is female, all LMs are biased (see \figureref{fig.anaphor.female.baseline}). The mono-lingual LMs strongly prefer a female object. 

On one hand because the LMs are strongly biased, using the female gender to test the anaphor phenomenon is inconclusive. Compare \figureref{fig.anaphor.female.self} to \figureref{fig.anaphor.female.baseline}, it is unclear whether the LMs achieved a high accuracy because they knew \textit{ziji} or just because they liked the female feature. The male self paradigm, on the other hand, shows that most mono-lingual LMs were able to use \textit{ziji} as a hint to agree the gender of the subject and object. Among the multi-lingual LMs, only \model{gpt3-davinci} achieved a meaningful accuracy increase.

Turning to the female self with \texttt{PP} paradigm in \figureref{fig.anaphor.female.pp}, even thought the mono-lingual LMs prefer the female feature in the baseline, when there is a male distractor in the \texttt{PP} which is linearly closer to the reflexive, the LMs are affected, reflected as a decrease in the accuracy. Fewer multi-lingual LMs are affected by the distractor. As a matter of fact, \model{XLM-large} and \model{ByT5-small} even have an increase in accuracy. On the male self with \texttt{PP} paradigm, only the \model{mengzi} models and \model{gpt3-davinci} are relatively unaffected by the distractor.

\noindent\textbf{Anaphor (Number)} The plural number feature is used to elicit the anaphor agreement. The feature is imposed on the subject by using numeral + classifier or the plural marker \textit{men}, or both. The plural feature on the object reflexive is reflected by adding \textit{men} to it. As it turns out, the number feature is not a good choice because most LMs are strongly biased (see \tableref{tab:18SLNG}).

\noindent\textbf{Aspect} Compared to \textit{le}, \textit{guo} has a fixed position in a \texttt{VP} and cannot take a wide scope over the progressive marker \textit{zai}. The results show that the LMs performed better on the \textit{guo} paradigms than on \textit{le}. There is no obvious reason why CPM in \figureref{fig.zai_V_le_Obj} performs extremely bad. 

\noindent\textbf{Classifier-noun agreement} The first paradigm tested the LMs' knowledge of the relative order of a demonstrative and classifier. \figureref{fig.cl.dem} shows that, except for the \model{CPM}, \model{PanGu-$\alpha$}, \model{mt5}, and \model{ByT5} models, all LMs' accuracy are comparable to the human annotators. 

\begin{figure*}[!h]
\centering
\resizebox{\textwidth}{!}{%
\includegraphics{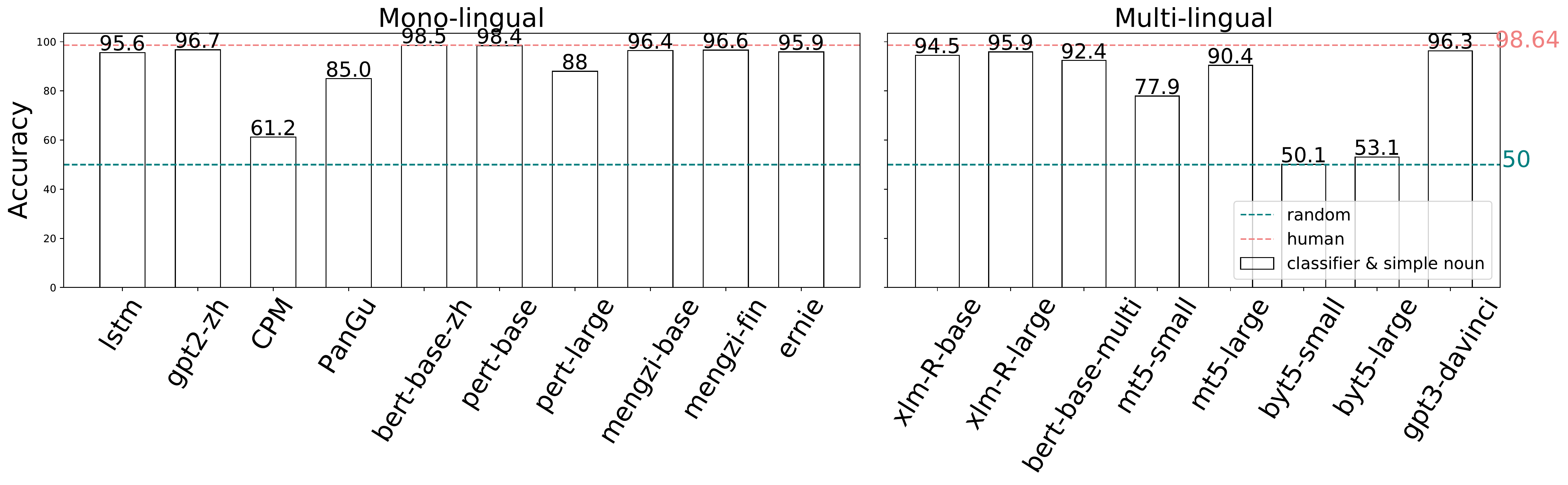}%
}
\caption{The LM accuracy on the classifier \& simple noun paradigm.}
\label{fig.cl.simple}
\end{figure*}

\begin{figure*}[!h]
\centering
\resizebox{\textwidth}{!}{%
\includegraphics{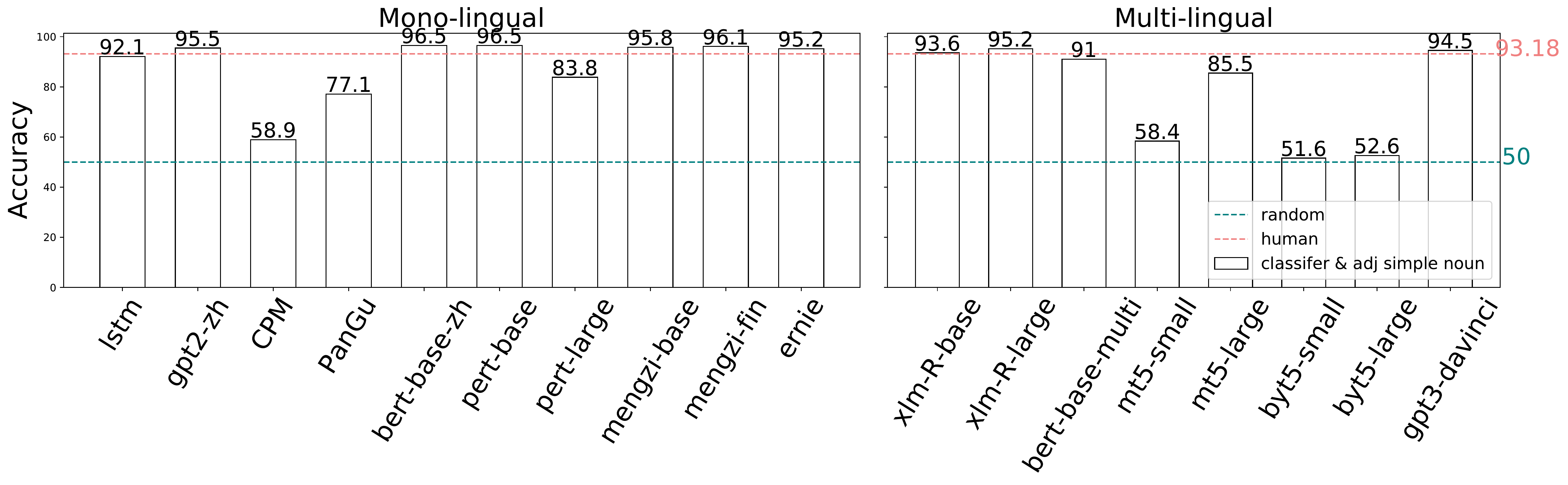}%
}
\caption{The LM accuracy on the classifier \& adj. simple noun paradigm.}
\label{fig.cl.simple.adj}
\end{figure*}

\begin{figure*}[!h]
\centering
\resizebox{\textwidth}{!}{%
\includegraphics{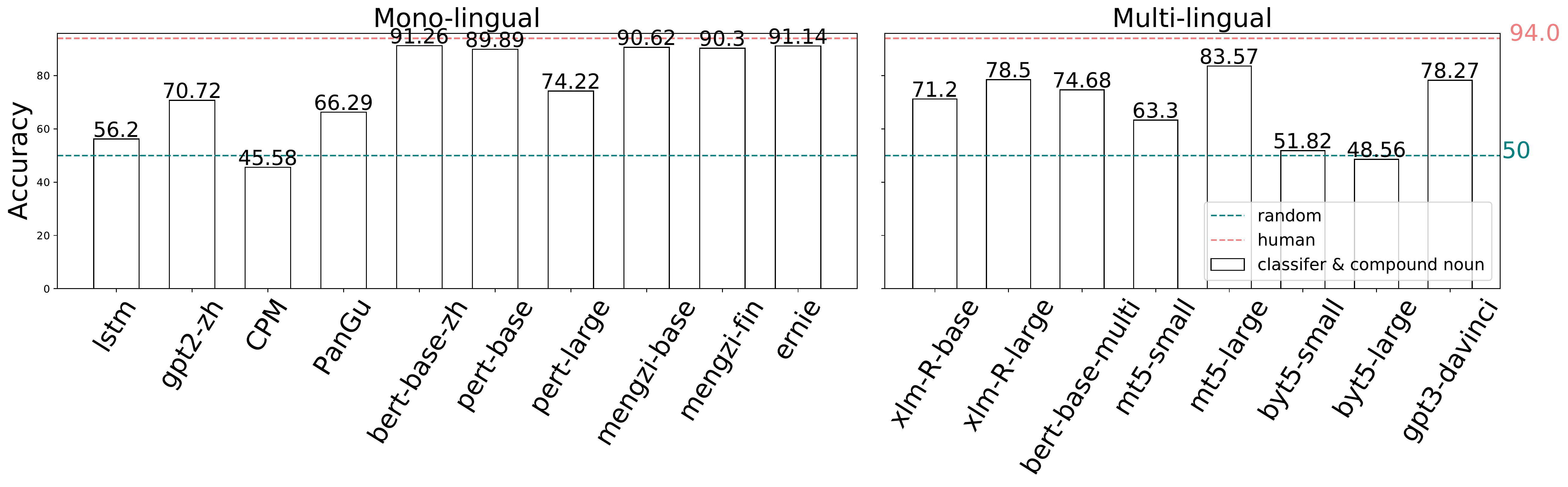}%
}
\caption{The LM accuracy on the classifier \& compound noun paradigm.}
\label{fig.cl.comp}
\end{figure*}

\begin{figure*}[!h]
\centering
\resizebox{\textwidth}{!}{%
\includegraphics{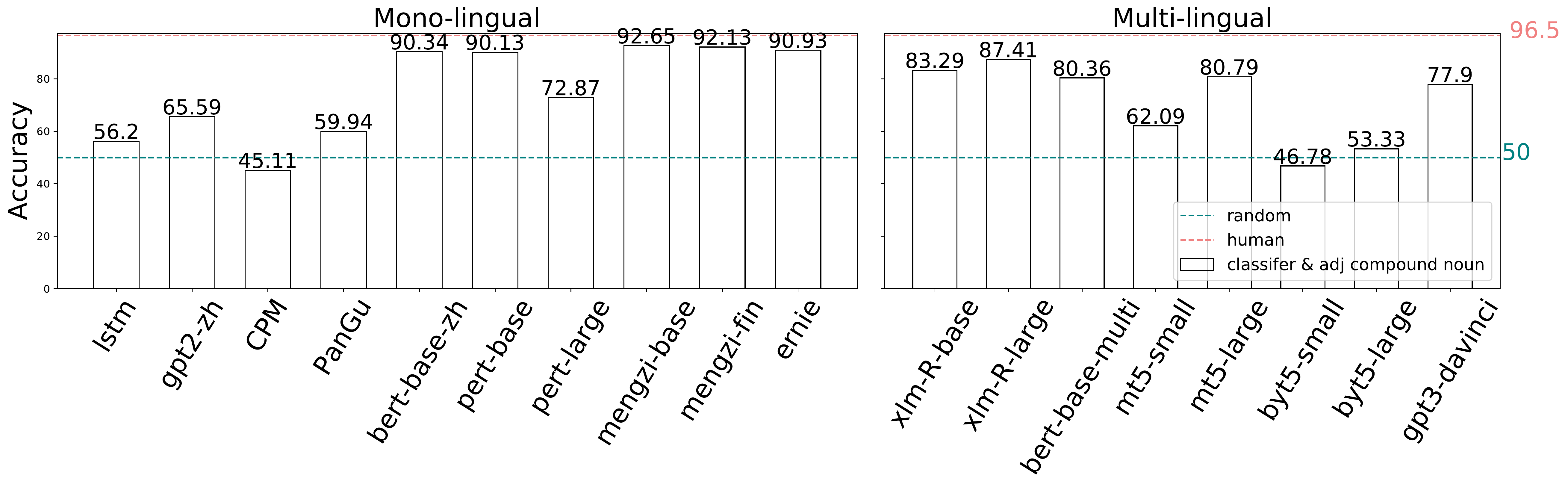}%
}
\caption{The LM accuracy on the classifier \& adj compound noun paradigm.}
\label{fig.cl.comp.adj}
\end{figure*}

Comparing the paradigms with simple nouns (\figureref{fig.cl.simple} and \ref{fig.cl.simple.adj}) to the ones with compound nouns (\figureref{fig.cl.comp} and \ref{fig.cl.comp.adj}), the multi-lingual models are more severely affected by the existence of a distractor (i.e., \texttt{noun1} in a compound noun) than the mono-lingual ones. The LMs are less affected by the distance created by the long adjective (\figureref{fig.cl.simple} \textit{vs.} \figureref{fig.cl.simple.adj}, and \figureref{fig.cl.comp} \textit{vs.} \figureref{fig.cl.comp.adj}).

\noindent\textbf{Definiteness Effect} Except for \model{CPM}, \model{PanGu-$\alpha$} and \model{pert-large}, all mono-lingual models have a decent accuracy. On the multi-lingual side, the \model{ByT5} models are especially bad. 

\noindent\textbf{Polarity item} Among the three PIs, \textit{huoduo huoshao} (more or less) reliably occurs only in affirmative contexts. The negative PIs, \textit{renhe} (any) and \textit{shenme} (what), can occur in negative, interrogative, and affirmative contexts. Fifteen out of eighteen LMs reached an accuracy on \textit{huoduo huoshao} comparable or even better than human. On the other two PIs, although there are quite a few LMs perform even better than human, overall, the accuracy values are worse and uneven. 

\noindent\textbf{Relative clause} In the resumptive noun paradigm, only \model{CPM} and \model{pert-large} have a satisfying performance. The other models are either near chance (\model{lstm} and \model{mt5-small}) or strongly deviated by the repeated filler in the gap position. The reason could be that the LMs are vulnerable to repetition, or to local grammaticality. When the gap in the relative clause is filled by a pronoun that matches the gender of the head noun, fewer than half of the LMs are able to notice the minimal pair contrast. 

\noindent\textbf{\textit{Wh}-fronting} All mono-lingual models performed well. Probably because \textit{wh} in situ is a prominent feature of Mandarin Chinese. Except for the \model{mt5} and \model{ByT5} models, most multi-lingual models did well. The \model{gpt3-davinci} model even reaches a $100$\% accuracy.


\section{Results}\label{appen.results}

\subsection{CLiMP}\label{appen.results.climp}

\begin{figure*}[!t]
\centering
\resizebox{\textwidth}{!}{%
\includegraphics{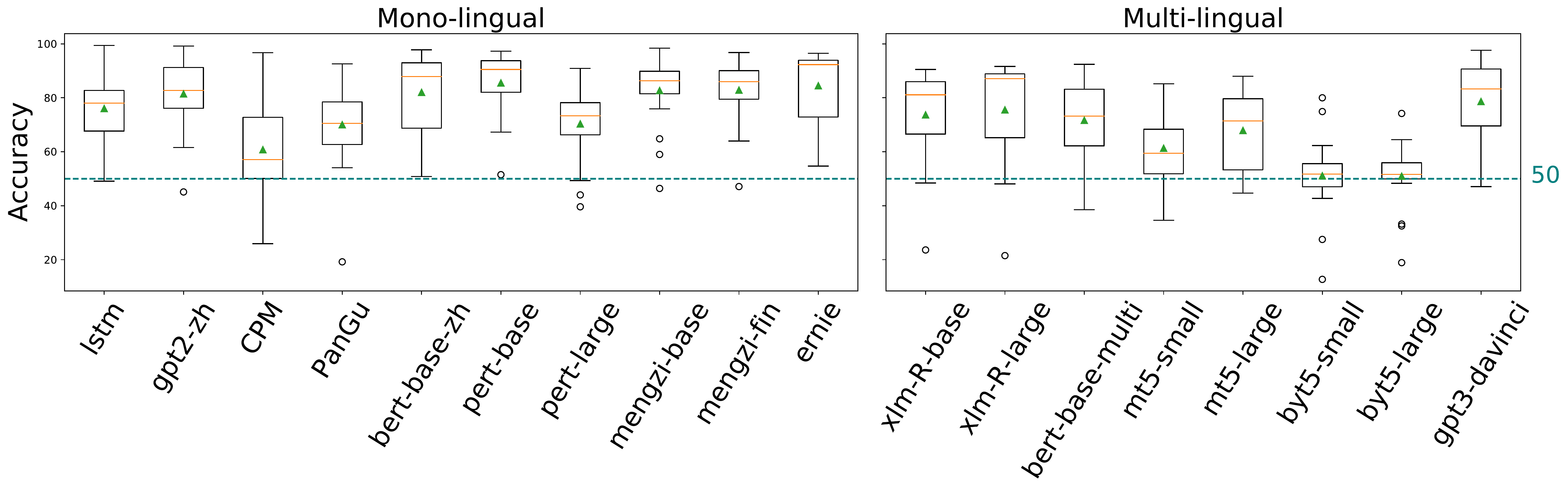}%
}
\caption{The box represents the inter-quartile range of the human and LM accuracy, with an orange line at the median accuracy and a green triangle at the mean. The whiskers extend from the box by 1.5 times. Dots are the accuracy values that past the end of the whiskers.}
\label{fig.all_lm_boxplot.climp}
\end{figure*}

The results are reported in \tableref{tab:17CLiMP} and \figureref{fig.all_lm_boxplot.climp}.


\begin{table*}[!h]
\centering
\resizebox{\textwidth}{!}{%
\begin{tabular}{@{}lrrrrrrrrrrrrrrrrrr@{}}
 &
  \rotatebox{70}{lstm} &
  \rotatebox{70}{gpt2-zh} &
  \rotatebox{70}{CPM} &
  \rotatebox{70}{PanGu} &
  \rotatebox{70}{bert-base-zh} &
  \rotatebox{70}{pert-base} &
  \rotatebox{70}{pert-large} &
  \rotatebox{70}{mengzi-base} &
  \rotatebox{70}{mengzi-base-fin} &
  \rotatebox{70}{ernie} &
  \rotatebox{70}{xlm-R-base} &
  \rotatebox{70}{xlm-R-large} &
  \rotatebox{70}{bert-base-multi} &
  \rotatebox{70}{mt5-small} &
  \rotatebox{70}{mt5-large} &
  \rotatebox{70}{byt5-small} &
  \rotatebox{70}{byt5-large} &
  \rotatebox{70}{gpt3} \\\toprule
anaphor\_agreement\_gender\_1000      & 82.6 & 79.5 & 79.9 & 92.6 & 86.2 & 90.5 & 71.1 & 96.1 & 96.2 & 93.7 & 82.1 & 78.0   & 73.0   & 46.2 & 69.3 & 55.4 & 49.4 & 83.3 \\
binding\_gender\_1000.csv             & 49.1 & 45.1 & 51.3 & 61.2 & 50.8 & 51.5 & 39.6 & 64.8 & 64.0   & 54.7 & 48.4 & 50.6 & 44.4 & 51.7 & 44.7 & 51.7 & 51.6 & 47.1 \\\midrule
ba\_construction\_1000                & 51.2 & 72.0   & 59.3 & 19.2 & 69.0   & 69.1 & 73.3 & 59.0   & 68.0   & 70.4 & 73.3 & 71.1 & 55.4 & 34.6 & 49.3 & 80.0   & 64.5 & 70.9 \\\midrule
classifier\_1000.csv                  & 90.8 & 95.1 & 57.1 & 76.0   & 95.6 & 95.4 & 78.8 & 89.3 & 90.2 & 96.5 & 85.6 & 90.8 & 87.8 & 58.6 & 77.4 & 49.9 & 51.7 & 93.1\\
classifier\_adj\_1000.csv             & 80.3 & 91.9 & 55.5 & 69.1 & 93.2 & 94.3 & 76.9 & 90.4 & 90.7 & 95.8 & 81.1 & 88.0   & 84.7 & 58.4 & 74.1 & 50.6 & 50.7 & 88.3 \\
classifier\_clause\_1000.csv          & 71.9 & 84.6 & 52.2 & 66.5 & 90.0   & 93.2 & 77.4 & 86.3 & 85.4 & 92.6 & 77.7 & 83.2 & 81.7 & 61.4 & 70.9 & 49.9 & 51.2 & 97.6 \\\midrule
coverb\_instrument\_1000.csv          & 62.7 & 82.7 & 36.0   & 54.1 & 91.1 & 97.3 & 63.9 & 92.6 & 93.8 & 96.3 & 89.3 & 90.4 & 60.0   & 52.0   & 80.7 & 54.9 & 55.7 & 87.6 \\
coverb\_with\_1000.csv                & 78.0   & 78.3 & 61.7 & 73.5 & 84.7 & 88.6 & 73.3 & 88.6 & 86.0   & 88.5 & 85.0   & 88.3 & 76.7 & 81.8 & 82.8 & 56.7 & 48.3 & 84.7 \\\midrule
filler\_gap\_dependency\_1000.csv     & 79.1 & 86.7 & 62.3 & 91.9 & 62.4 & 80.2 & 90.9 & 86.3 & 82.7 & 70.1 & 67.9 & 60.3 & 78.2 & 80.3 & 46.0   & 62.3 & 63.3 & 68.2 \\\midrule
head\_final\_clause\_1000.csv         & 68.3 & 77.0   & 86.5 & 65.6 & 53.1 & 83.9 & 73.3 & 82.5 & 78.9 & 78.0   & 76.2 & 87.1 & 72.0   & 85.2 & 85.8 & 43.6 & 60.6 & 73.0 \\\midrule
passive\_formal\_1000.csv             & 69.2 & 61.6 & 47.0   & 61.6 & 67.7 & 67.3 & 44.0   & 46.4 & 47.1 & 68.7 & 55.0   & 48.1 & 73.2 & 57.3 & 51.4 & 54.2 & 52.4 & 54.5 \\\midrule
verb\_complement\_direction\_1000.csv & 67.0   & 75.2 & 81.4 & 80.1 & 93.0   & 91.4 & 85.9 & 83.3 & 89.2 & 71.6 & 90.5 & 88.4 & 38.5 & 50.7 & 55.2 & 42.7 & 56.1 & 73.2 \\
verb\_complement\_duration\_1000.csv  & 96.1 & 99.1 & 83.6 & 82.6 & 90.2 & 96.4 & 89.1 & 98.4 & 96.8 & 94.1 & 86.4 & 90.4 & 76.3 & 64.6 & 51.0   & 12.7 & 18.9 & 55.4\\
verb\_complement\_frequency\_1000.csv & 98.5 & 99.2 & 48.8 & 75.6 & 97.8 & 91.5 & 78.7 & 75.9 & 75.0   & 87.5 & 23.6 & 21.5 & 90.9 & 69.8 & 71.4 & 44.2 & 32.5 & 96.0 \\
verb\_complement\_res\_adj\_1000.csv  & 82.9 & 87.5 & 25.9 & 59.3 & 87.6 & 87.0   & 49.3 & 85.5 & 84.2 & 92.5 & 90.2 & 91.6 & 64.4 & 71.9 & 88.0   & 74.9 & 74.2 & 79.3 \\
verb\_complement\_res\_verb\_1000.csv & 99.4 & 98.5 & 96.7 & 90.1 & 96.2 & 88.8 & 68.9 & 85.9 & 87.2 & 92.3 & 53.6 & 66.1 & 92.4 & 65.0   & 78.6 & 27.5 & 33.2 & 97.0 \\\midrule
Average over 8 phenomena & 71.7 & 77.8 & 61.5 & 65.9 & 74.3 & 81.2 & 69.7 & 77.5 & 77.7 & 79.6 & 71.9 & 72.4 & 70.4 & 62.1 & 64.3 & 55.0 & 54.7 & 73.9\\
Std-dev over 8 phenomena & 11.4 & 11.9	& 12.5	& 21.2	& 15.0	& 11.1	& 14.3 & 16.0 & 14.2 & 10.6 & 10.0 & 14.8 & 9.6 & 16.3 & 15.4 & 12.2 & 7.4 & 12.2 \\
\bottomrule
\end{tabular}}
\caption{Eighteen LMs' performance on CLiMP.}
\label{tab:17CLiMP}
\end{table*}

\subsection{SLING}\label{appen.results.sling}

\begin{figure*}[!t]
\centering
\resizebox{\textwidth}{!}{%
\includegraphics{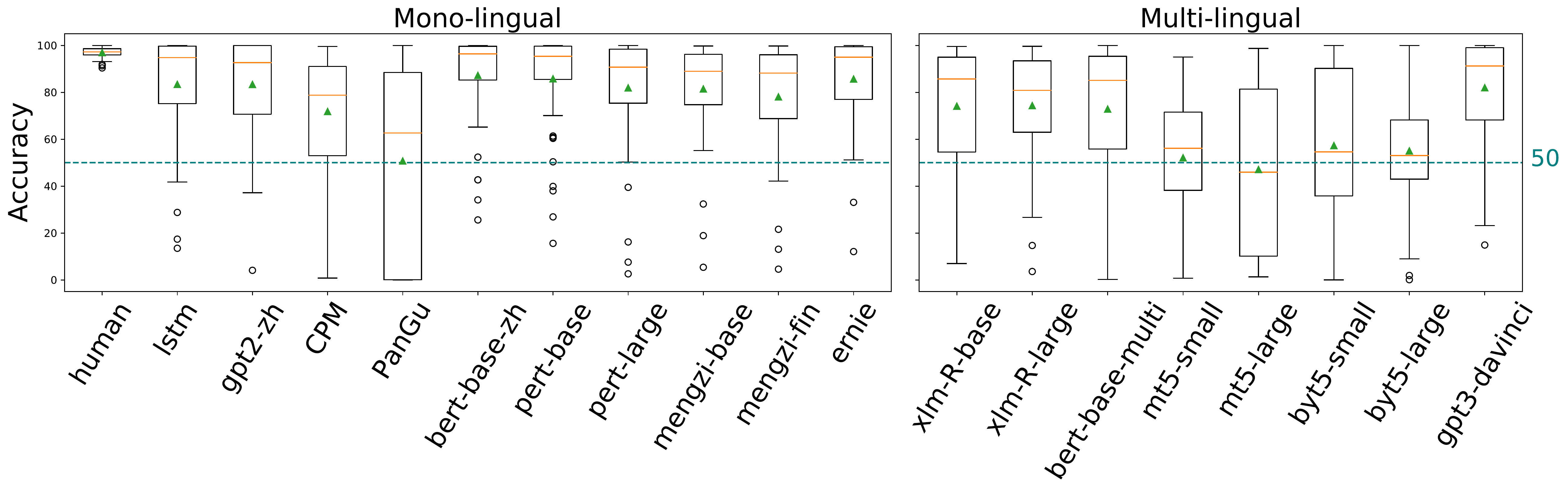}%
}
\caption{The box represents the inter-quartile range of the human and LM accuracy, with an orange line at the median accuracy and a green triangle at the mean. The whiskers extend from the box by 1.5 times. Dots are the accuracy values that past the end of the whiskers.}
\label{fig.all_lm_boxplot}
\end{figure*}

\begin{figure*}[!h]
\centering
\resizebox{\textwidth}{!}{%
\includegraphics{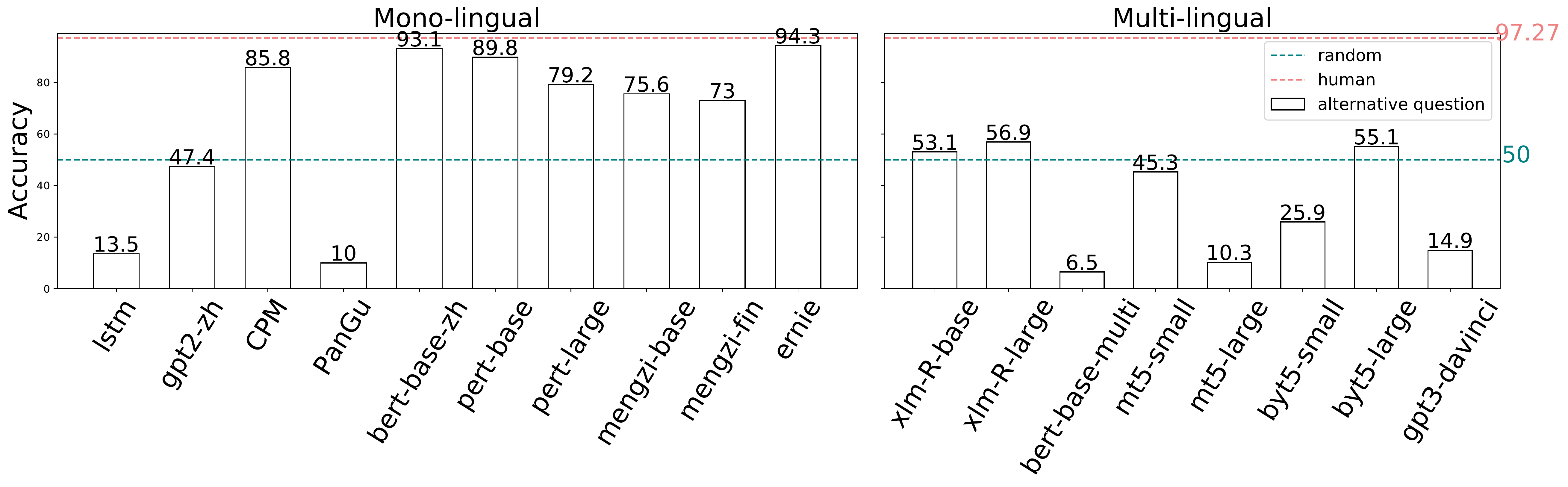}%
}
\caption{The LM accuracy on the alternative question phenomenon.}
\label{fig.altq}
\end{figure*}

\begin{figure*}[!h]
\centering
\resizebox{\textwidth}{!}{%
\includegraphics{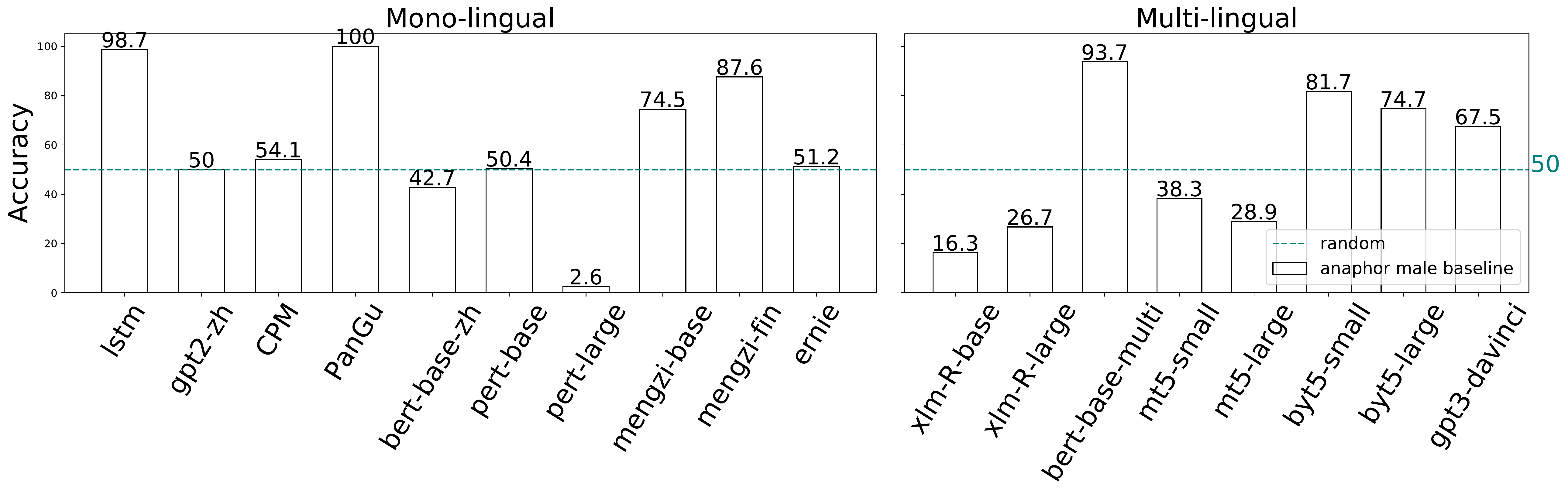}%
}
\caption{The LM bias towards a male object when the subject is male.}
\label{fig.anaphor.male.baseline}
\end{figure*}

\begin{figure*}[!h]
\centering
\resizebox{\textwidth}{!}{%
\includegraphics{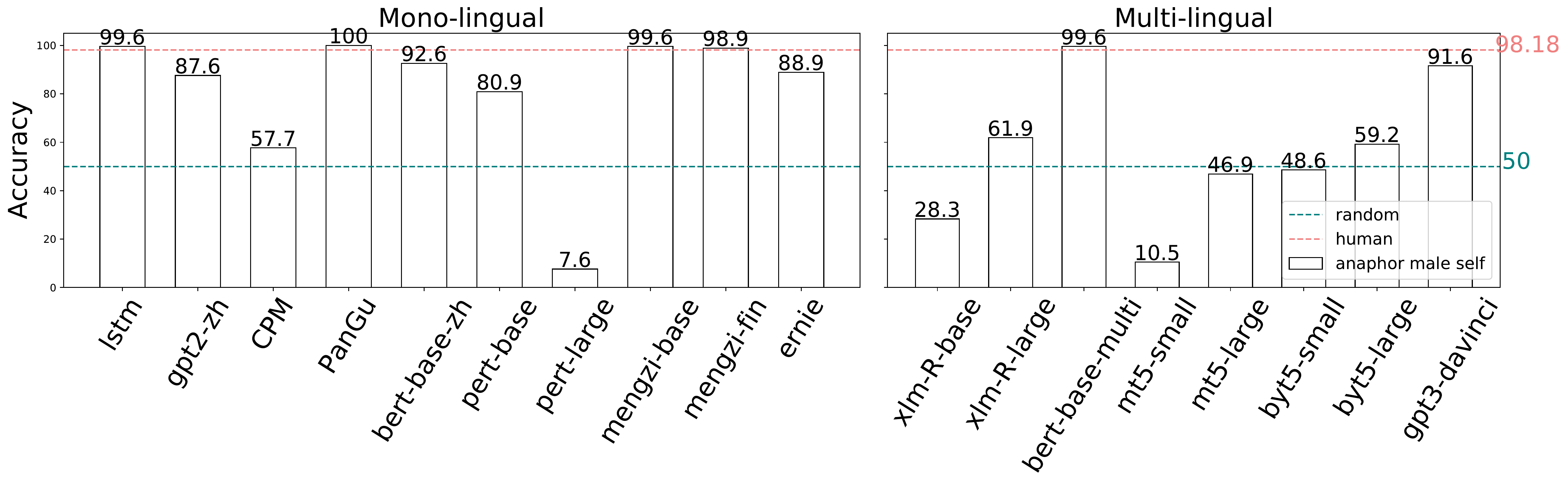}%
}
\caption{The LM accuracy on the anaphor male self paradigm.}
\label{fig.anaphor.male.self}
\end{figure*}

\begin{figure*}[!h]
\centering
\resizebox{\textwidth}{!}{%
\includegraphics{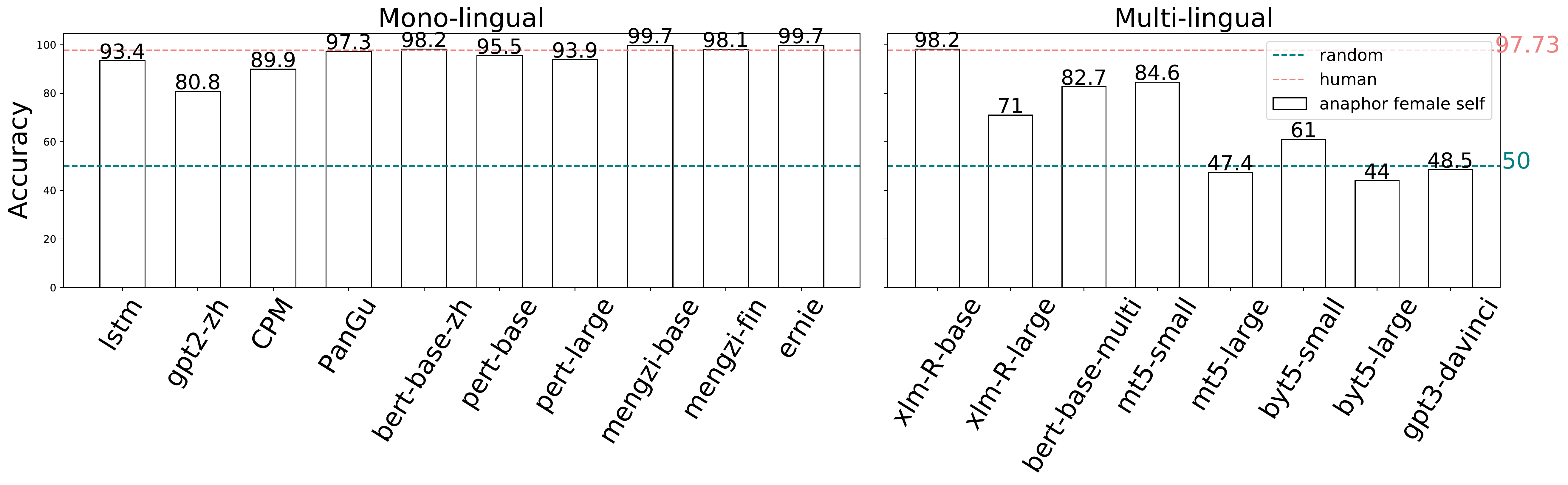}%
}
\caption{The LM accuracy on the anaphor female self paradigm.}
\label{fig.anaphor.female.self}
\end{figure*}

\begin{figure*}[!h]
\centering
\resizebox{\textwidth}{!}{%
\includegraphics{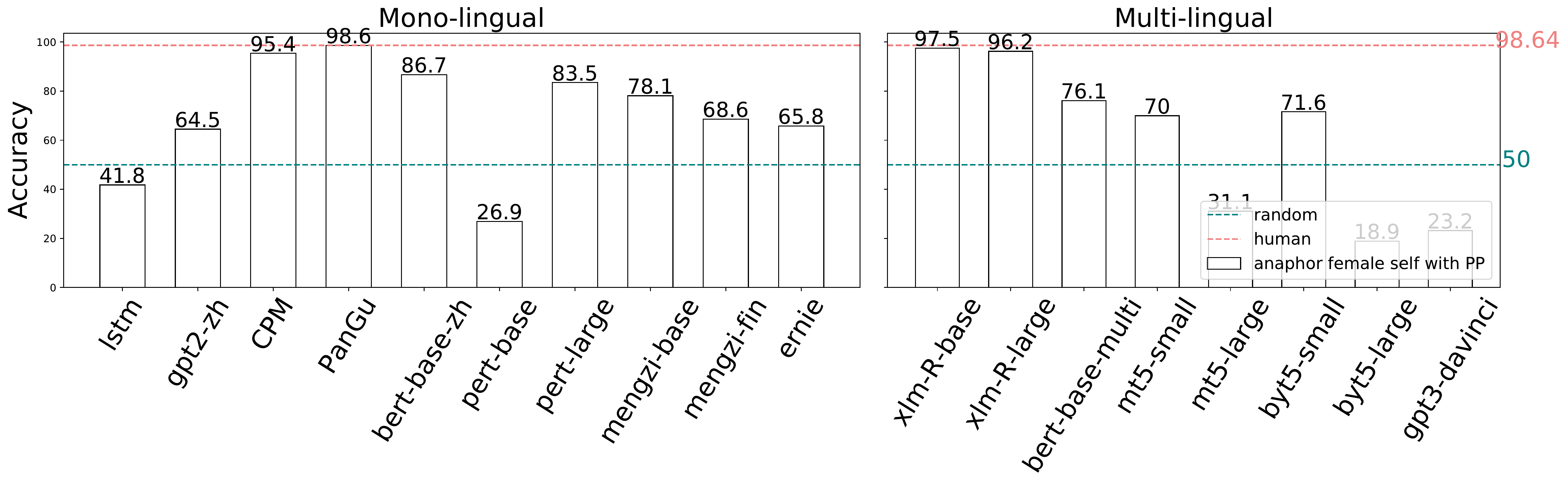}%
}
\caption{The LM accuracy on the anaphor female self with PP paradigm.}
\label{fig.anaphor.female.pp}
\end{figure*}

\begin{figure*}[!h]
\centering
\resizebox{\textwidth}{!}{%
\includegraphics{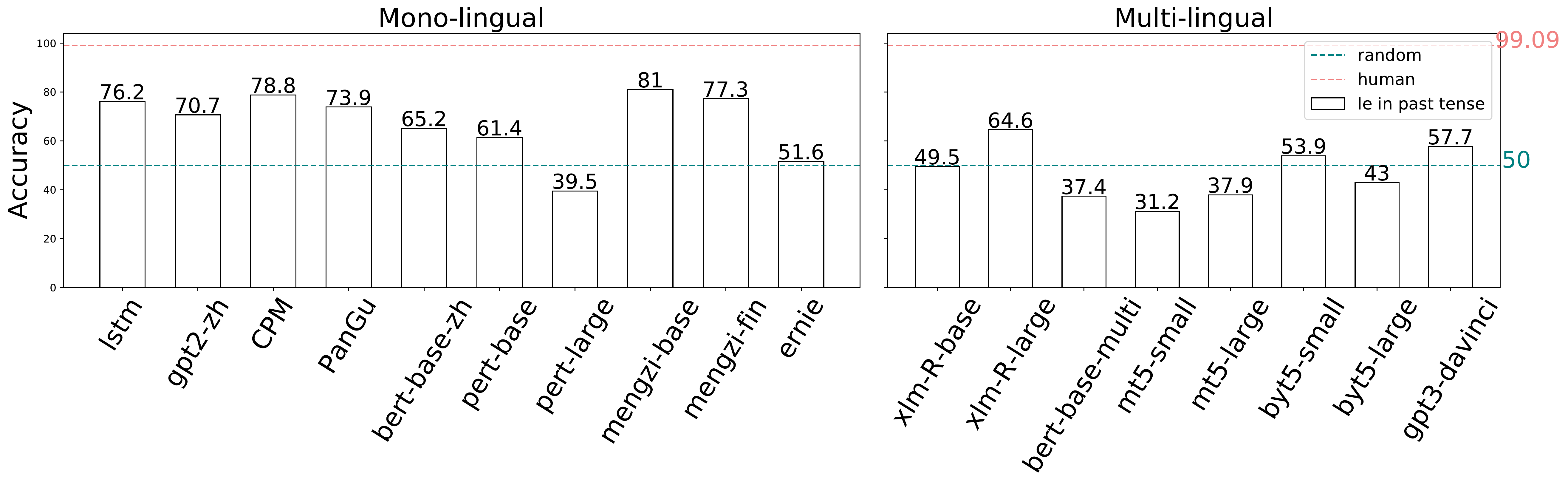}%
}
\caption{The LM accuracy on the le in past tense paradigm.}
\label{fig.aspect.le.past}
\end{figure*}

\begin{figure*}[!h]
\centering
\resizebox{\textwidth}{!}{%
\includegraphics{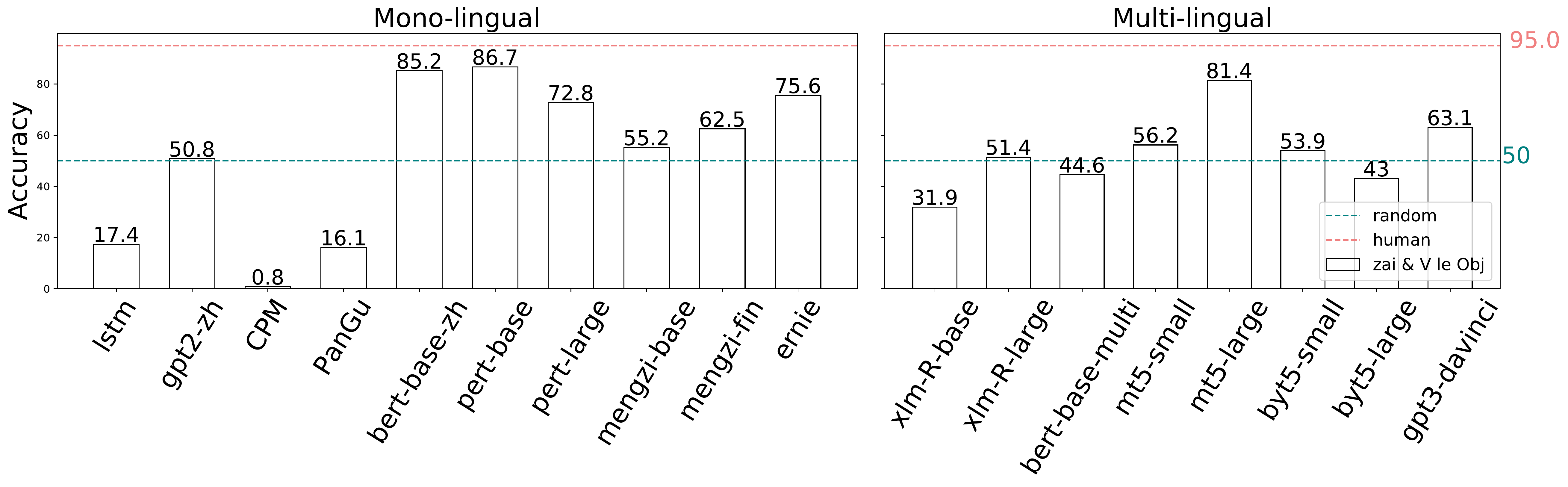}%
}
\caption{The LM accuracy on the zai \& V le Obj paradigm.}
\label{fig.zai_V_le_Obj}
\end{figure*}

\begin{figure*}[!h]
\centering
\resizebox{\textwidth}{!}{%
\includegraphics{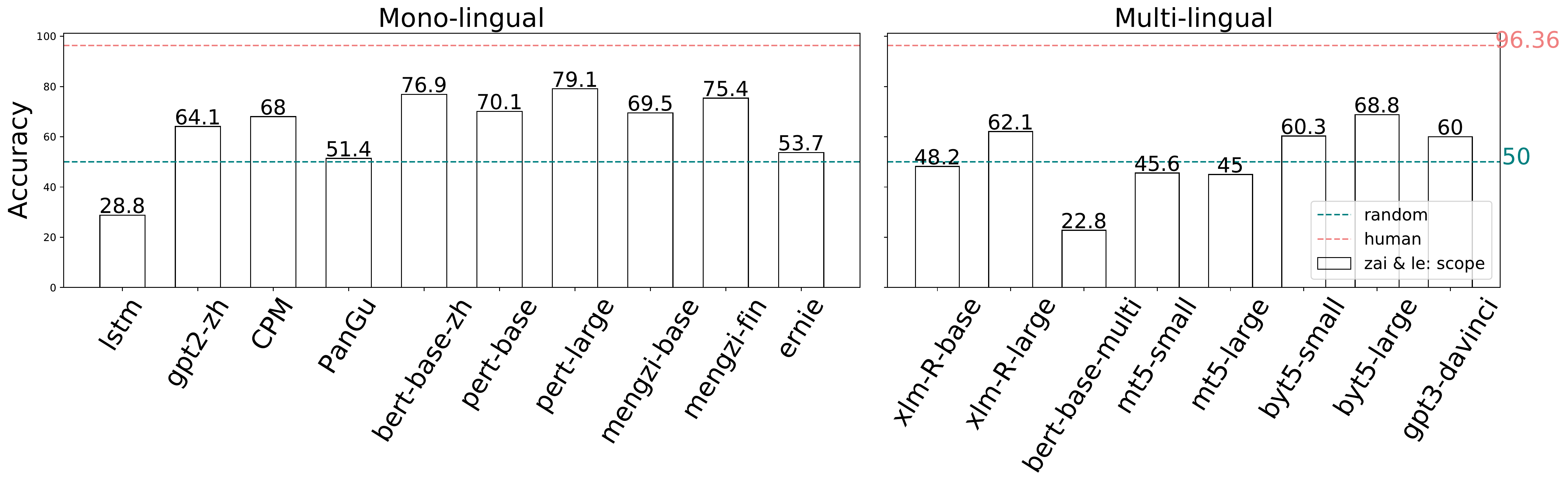}%
}
\caption{The LM accuracy on the zai \& le scope paradigm.}
\label{fig.zai_le_scope}
\end{figure*}

\begin{figure*}[!h]
\centering
\resizebox{\textwidth}{!}{%
\includegraphics{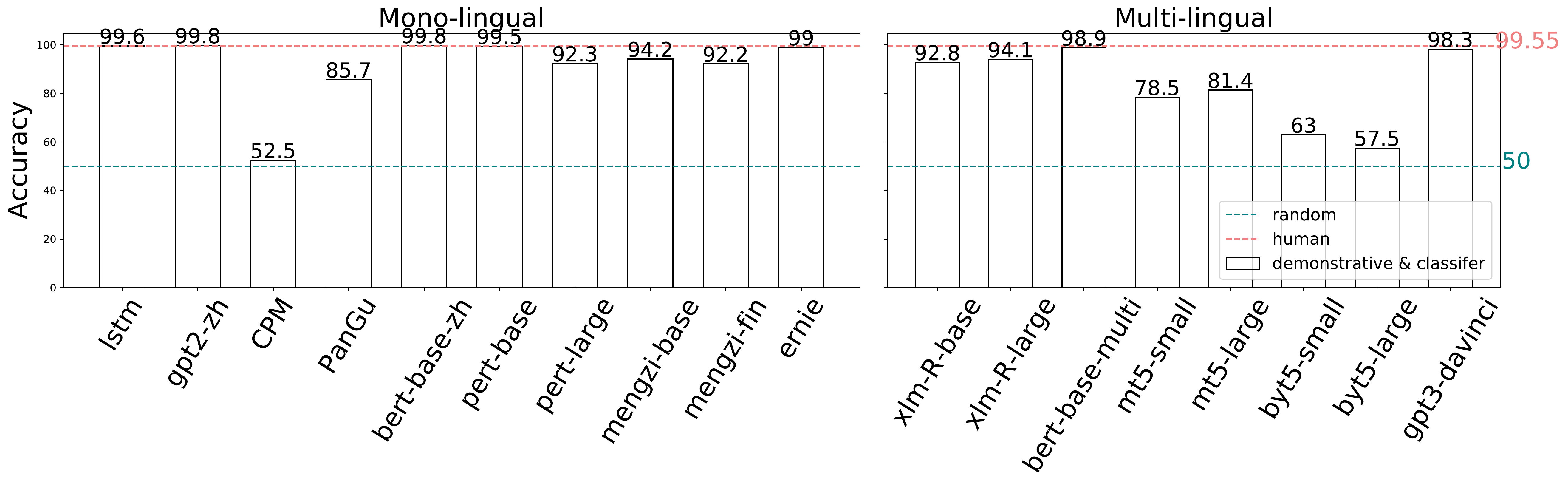}%
}
\caption{The LM accuracy on the demonstrative \& classifier paradigm.}
\label{fig.cl.dem}
\end{figure*}

\begin{figure*}[!h]
\centering
\resizebox{\textwidth}{!}{%
\includegraphics{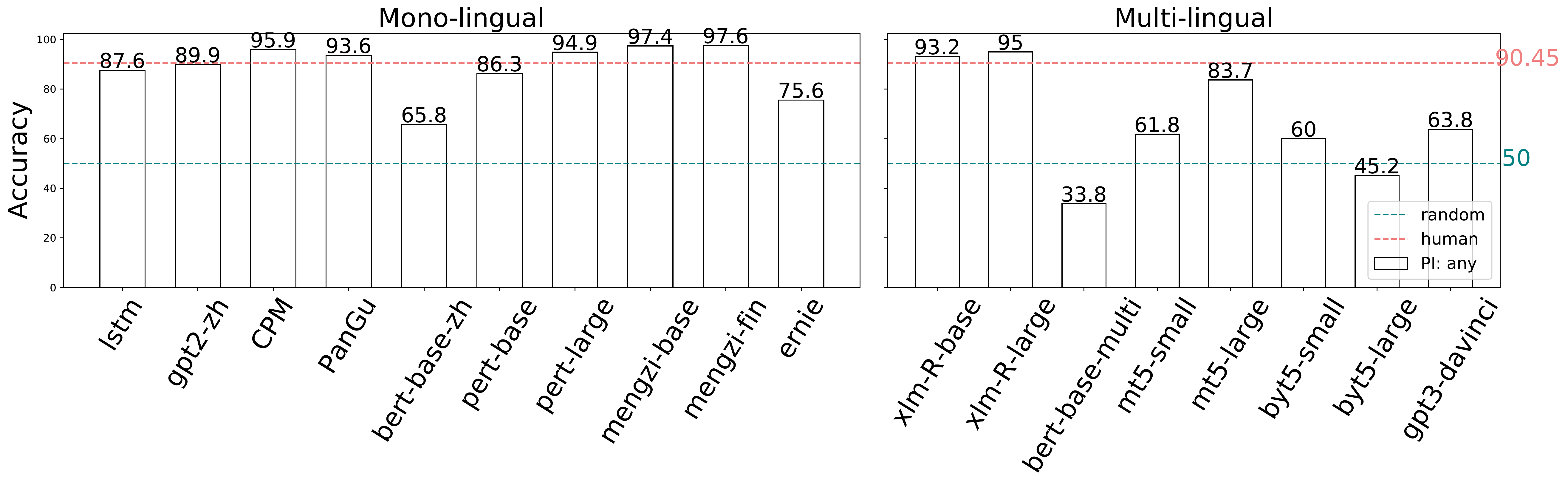}%
}
\caption{The LM accuracy on the polarity item any paradigm.}
\label{fig.any}
\end{figure*}

\begin{figure*}[!h]
\centering
\resizebox{\textwidth}{!}{%
\includegraphics{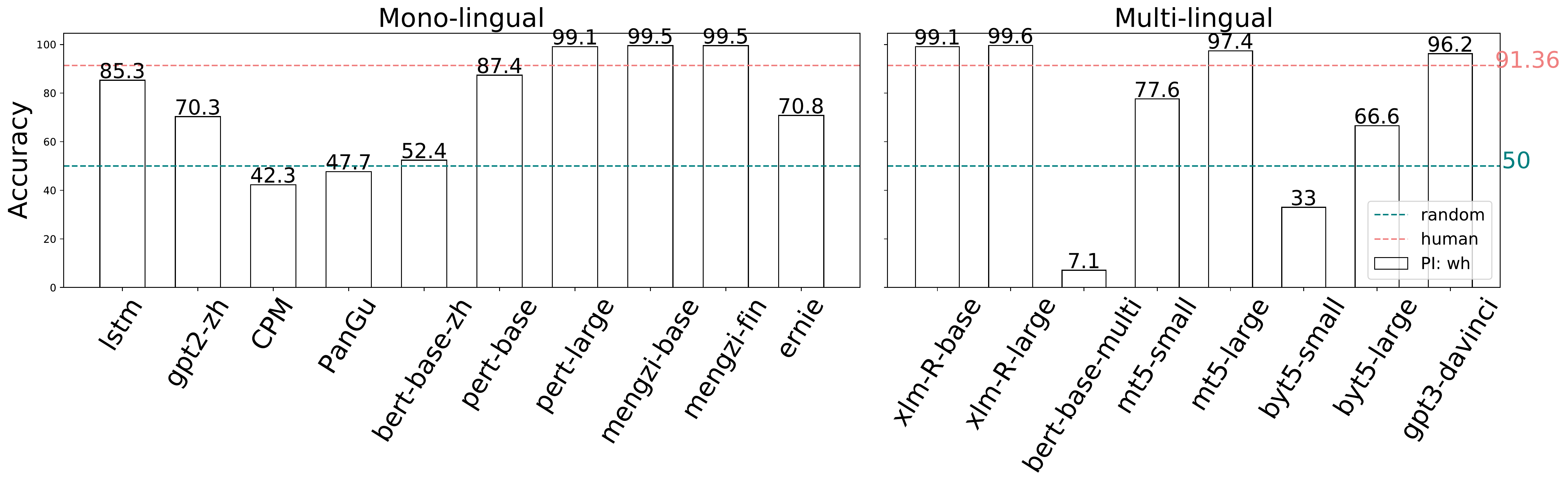}%
}
\caption{The LM accuracy on the polarity item wh paradigm.}
\label{fig.PI_wh}
\end{figure*}

\begin{figure*}[!h]
\centering
\resizebox{\textwidth}{!}{%
\includegraphics{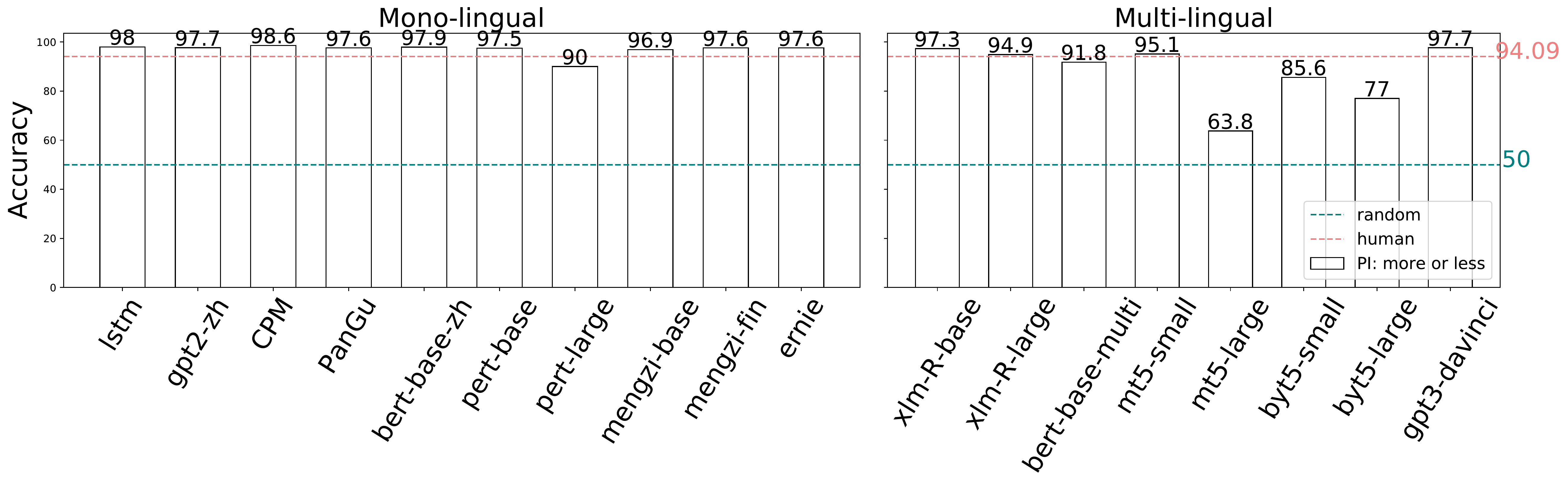}%
}
\caption{The LM accuracy on the polarity item more or less paradigm.}
\label{fig.moreorless}
\end{figure*}

\begin{figure*}[!h]
\centering
\resizebox{\textwidth}{!}{%
\includegraphics{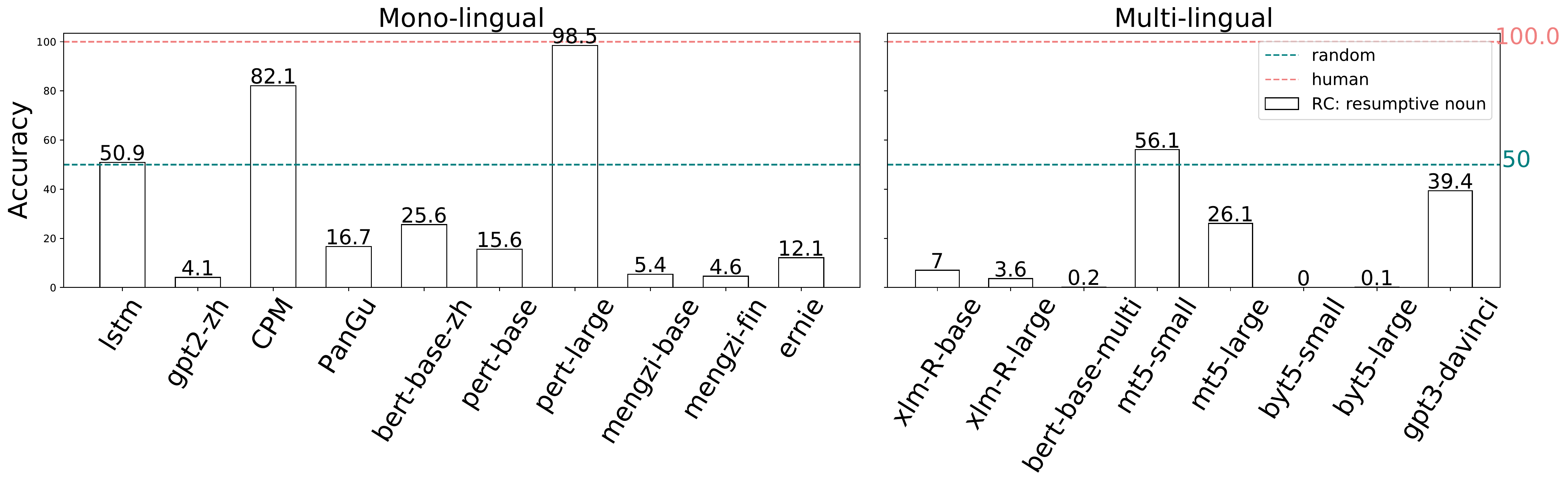}%
}
\caption{The LM accuracy on the relative clause with resumptive noun paradigm.}
\label{fig.resumptive.noun}
\end{figure*}

\begin{figure*}[!h]
\centering
\resizebox{\textwidth}{!}{%
\includegraphics{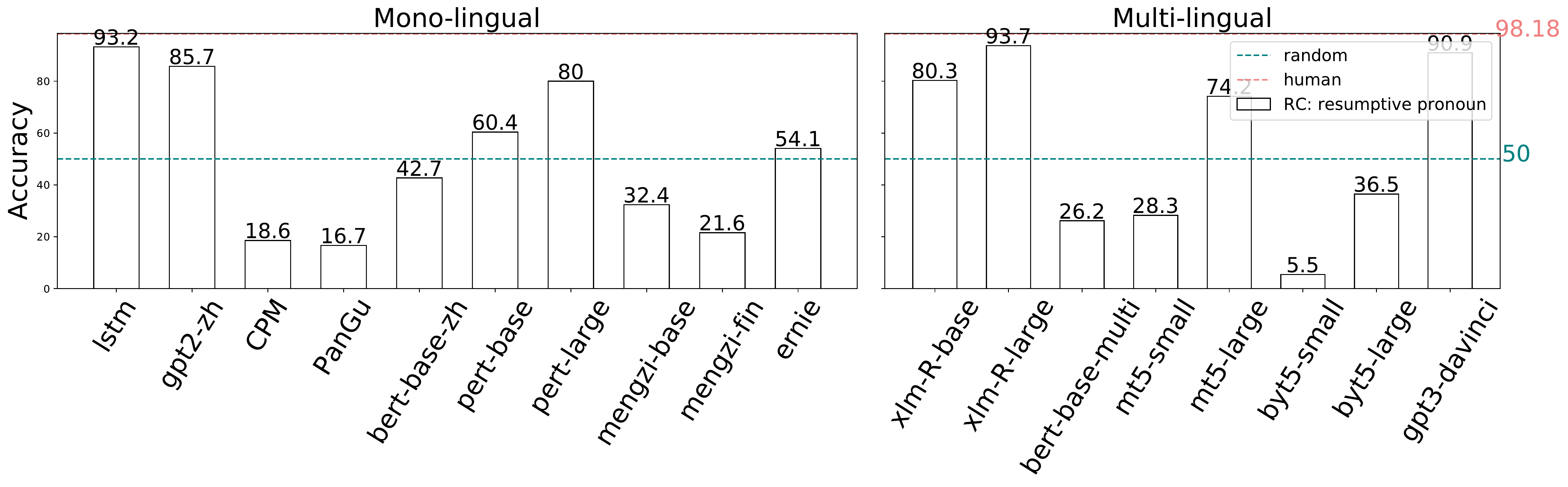}%
}
\caption{The LM accuracy on the relative clause with resumptive pronoun paradigm.}
\label{fig.resumptive.pronoun}
\end{figure*}

\begin{figure*}[!h]
\centering
\resizebox{\textwidth}{!}{%
\includegraphics{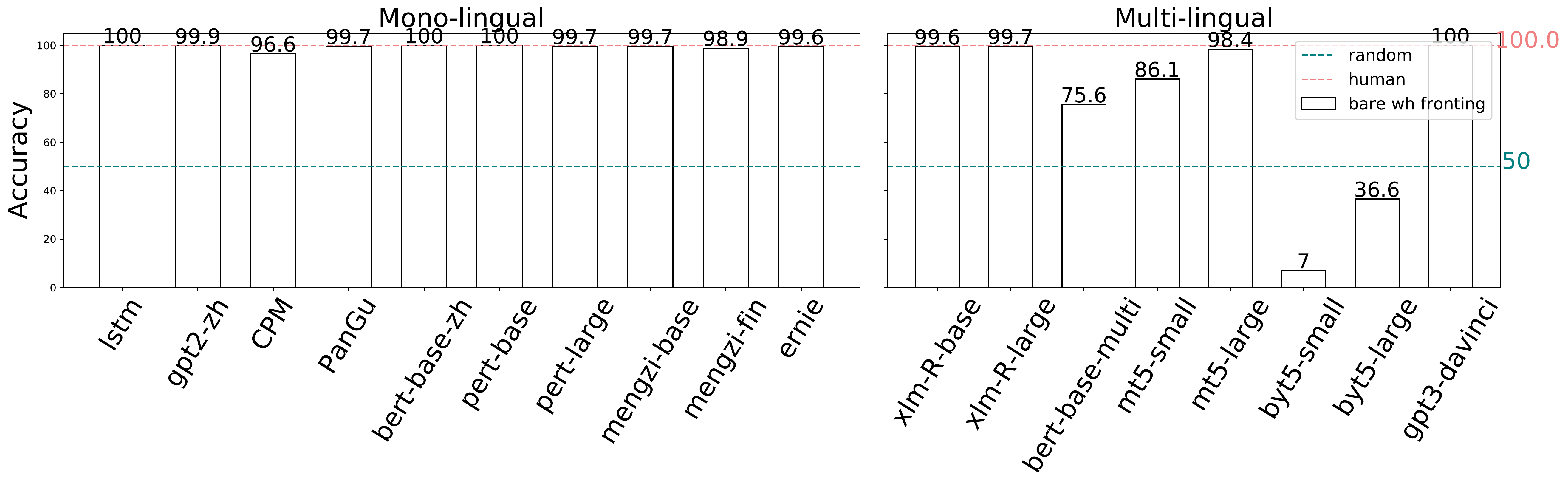}%
}
\caption{The LM accuracy on the bare wh fronting paradigm.}
\label{fig.barewh}
\end{figure*}

\begin{figure*}[!h]
\centering
\resizebox{\textwidth}{!}{%
\includegraphics{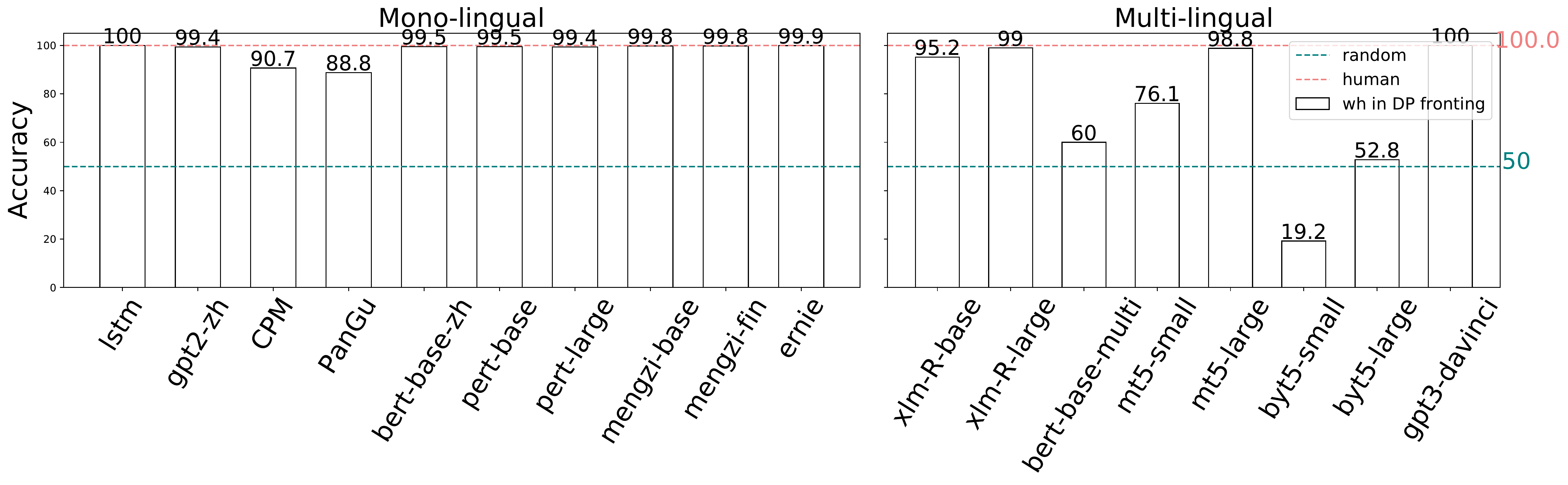}%
}
\caption{The LM accuracy on the wh in DP fronting paradigm.}
\label{fig.whasmodifier}
\end{figure*}

The results are reported in \tableref{tab:18SLNG} and \figureref{fig.all_lm_boxplot} to \figureref{fig.whasmodifier}.

\begin{table*}[!h]
\centering
\resizebox{\textwidth}{!}{%
\begin{tabular}{@{}ll>{\bf}rrrrrrrrrrrrrrrrrrr@{}}
 &
   &
  \rotatebox{70}{human} &
  \rotatebox{70}{lstm} &
  \rotatebox{70}{gpt2-zh} &
  \rotatebox{70}{CPM} &
  \rotatebox{70}{PanGu} &
  \rotatebox{70}{bert-base-zh} &
  \rotatebox{70}{pert-base} &
  \rotatebox{70}{pert-large} &
  \rotatebox{70}{mengzi-base} &
  \rotatebox{70}{mengzi-base-fin} &
  \rotatebox{70}{ernie} &
  \rotatebox{70}{xlm-R-base} &
  \rotatebox{70}{xlm-R-large} &
  \rotatebox{70}{bert-base-multi} &
  \rotatebox{70}{mt5-small} &
  \rotatebox{70}{mt5-large} &
  \rotatebox{70}{byt5-small} &
  \rotatebox{70}{byt5-large} &
  \rotatebox{70}{gpt3} \\\toprule
Alternative question & haishi & 97.3 & 13.5 & 47.4 & 85.8 & 10.0 & 93.1 & 89.8 & 79.2 & 75.6 & 73.0 & 94.3 & 53.1 & 56.9 & 6.5 & 45.3 & 10.3 & 25.9 & 55.1 & 14.9 \\\midrule
\multirow{6}{*}{Anaphor (gender)} &\cellcolor{aliceblue}
 male\_baseline & \cellcolor{aliceblue}  & \cellcolor{aliceblue} 98.7 & \cellcolor{aliceblue} 50.0 & \cellcolor{aliceblue} 54.1 & \cellcolor{aliceblue} 100 & \cellcolor{aliceblue} 42.7 & \cellcolor{aliceblue} 50.4 & \cellcolor{aliceblue} 2.6 & \cellcolor{aliceblue} 74.5 & \cellcolor{aliceblue} 87.6 & \cellcolor{aliceblue} 51.2 &  \cellcolor{aliceblue} 16.3 & \cellcolor{aliceblue} 26.7 & \cellcolor{aliceblue} 93.7 &  \cellcolor{aliceblue} 38.3 & \cellcolor{aliceblue} 28.9 & \cellcolor{aliceblue} 81.7 &  \cellcolor{aliceblue} 74.7 &  \cellcolor{aliceblue} 67.5 \\
 &
  male\_self &  98.2 &  99.6 &  87.6 &  57.7 &  100 &  92.6 &  80.9 &  7.6 &  99.6 &  98.9 &  88.9 &  28.3 &  61.9 &  99.6 &  10.5 &  46.9 &  48.6 &  59.2 &  91.6 \\
 &
  pp\_male & 99.6 & 64.9 & 37.2 & 41.3 & 99.9 & 75.7 & 39.9 & 16.2 & 91.4 & 91.6 & 71.8 & 13.9 & 14.7 & 71.5 & 37.4 & 25.3 & 34.3 & 28.7 & 89.6 \\
 &\cellcolor{aliceblue}
  female\_baseline & \cellcolor{aliceblue}
  &
  \cellcolor{aliceblue} 92.8 & \cellcolor{aliceblue} 80.2 & \cellcolor{aliceblue} 91.2 &  \cellcolor{aliceblue} 95.9 & \cellcolor{aliceblue} 85.6 & \cellcolor{aliceblue} 85.3 &  \cellcolor{aliceblue} 94.9 & \cellcolor{aliceblue} 89.3 & \cellcolor{aliceblue} 91.2 &  \cellcolor{aliceblue} 96.4 & \cellcolor{aliceblue} 90.5 & \cellcolor{aliceblue} 62.8 &  \cellcolor{aliceblue} 82.8 & \cellcolor{aliceblue} 64.0 &  \cellcolor{aliceblue} 39.2 &  \cellcolor{aliceblue} 40.6 & \cellcolor{aliceblue} 48.0 &  \cellcolor{aliceblue} 40.8 \\
 &
  female\_self &
  97.7 &  93.4 &  80.8 &  89.9 &  97.3 &  98.2 &  95.5 &  93.9 &  99.7 &  98.1 &  99.7 &  98.2 &  71.0 &  82.7 &  84.6 &  47.4 &  61.0 &  44.0 &  48.5 \\
 &
  pp\_female &  98.6 &  41.8 &  64.5 &  95.4 &  98.6 &  86.7 &  26.9 &  83.5 &  78.1 &  68.6 &  65.8 &  97.5 &  96.2 &  76.1 &  70.0 &  31.1 &  71.6 &  18.9 &  23.2 \\\midrule
\multirow{12}{*}{Anaphor (number)} &
\cellcolor{aliceblue}
  cl\_male\_baseline & \cellcolor{aliceblue}  & \cellcolor{aliceblue} 100 & \cellcolor{aliceblue} 100 & \cellcolor{aliceblue} 76.4 & \cellcolor{aliceblue} 0.0 & \cellcolor{aliceblue} 99.7 & \cellcolor{aliceblue} 100 & \cellcolor{aliceblue} 97.7 &  \cellcolor{aliceblue} 82.3 & \cellcolor{aliceblue} 67.8 & \cellcolor{aliceblue} 100 &  \cellcolor{aliceblue} 94.3 & \cellcolor{aliceblue} 81.7 & \cellcolor{aliceblue} 99.7 &  \cellcolor{aliceblue} 80.6 & \cellcolor{aliceblue} 7.0 & \cellcolor{aliceblue} 99.9 & \cellcolor{aliceblue} 99.5 & \cellcolor{aliceblue} 99.9 \\
 &
  cl\_self\_male &  95.5 &  99.5 &  100 &  98.7 &  0.1 &  99.7 &  100 &  99.3 &  78.5 &  69.4 &  99.8 &  98.5 &  88.1 &  97.3 &  68.1 &  5.0 &  99.4 &  61.7 &  99.8 \\
 &\cellcolor{aliceblue}
  cl\_female\_baseline & \cellcolor{aliceblue}  & \cellcolor{aliceblue} 99.5 & \cellcolor{aliceblue} 100 & \cellcolor{aliceblue} 80.6 & \cellcolor{aliceblue}0.0 & \cellcolor{aliceblue} 99.6 & \cellcolor{aliceblue} 99.9 & \cellcolor{aliceblue} 91.2 & \cellcolor{aliceblue} 80.0 & \cellcolor{aliceblue} 65.8 & \cellcolor{aliceblue} 99.2 & \cellcolor{aliceblue} 67.5 & \cellcolor{aliceblue} 42.3 & \cellcolor{aliceblue} 98.5 &  \cellcolor{aliceblue} 19.6 & \cellcolor{aliceblue} 5.9 & \cellcolor{aliceblue} 91.8 &  \cellcolor{aliceblue} 65.2 & \cellcolor{aliceblue} 99.0 \\
 &
  cl\_self\_female &  97.3 &  98.7 &  100 &  98.6 & 0.0 &  100 &  99.6 &  98.9 &  72.6 &  83.4 &  99.6 &  94.3 &  69.6 &  89.6 &  2.1 &  8.1 &  63.1 &  42.9 &  99.1 \\
 &\cellcolor{aliceblue}
  men\_male\_baseline & \cellcolor{aliceblue}  & \cellcolor{aliceblue} 100 & \cellcolor{aliceblue} 100 & \cellcolor{aliceblue} 48.4 & \cellcolor{aliceblue} 0.1 & \cellcolor{aliceblue} 99.8 & \cellcolor{aliceblue} 100 & \cellcolor{aliceblue} 98.3 &  \cellcolor{aliceblue} 66.3 & \cellcolor{aliceblue} 43.5 & \cellcolor{aliceblue} 99.6 &  \cellcolor{aliceblue} 84.4 & \cellcolor{aliceblue} 77.5 & \cellcolor{aliceblue} 100 &  \cellcolor{aliceblue} 45.1 & \cellcolor{aliceblue} 5.8 & \cellcolor{aliceblue} 98.7 & \cellcolor{aliceblue} 100 & \cellcolor{aliceblue} 99.8 \\
 &
  menself\_male &  97.3 &  100 &  100 &  79.9 & 0.0 &  99.9 &  100 &  100 &  94.1 &  85.8 &  99.7 &  98.4 &  85.2 &  99.5 &  52.3 &  5.2 &  100 &  77.4 &  100 \\
 &\cellcolor{aliceblue}
  men\_female\_baseline &  \cellcolor{aliceblue} & \cellcolor{aliceblue} 100 & \cellcolor{aliceblue} 100 & \cellcolor{aliceblue} 50.5 & \cellcolor{aliceblue}0.0 & \cellcolor{aliceblue} 98.0 & \cellcolor{aliceblue} 99.9 & \cellcolor{aliceblue} 96.6 & \cellcolor{aliceblue} 65.4 & \cellcolor{aliceblue} 48.1 & \cellcolor{aliceblue} 97.6 &  \cellcolor{aliceblue} 54.4 & \cellcolor{aliceblue} 63.6 & \cellcolor{aliceblue} 99.6 &  \cellcolor{aliceblue} 7.1 & \cellcolor{aliceblue} 10.1 & \cellcolor{aliceblue} 99.4 &  \cellcolor{aliceblue} 99.9 & \cellcolor{aliceblue} 96.3 \\
 &
  menself\_female &  95.9 &  100 &  100 &  80.9 & 0.0 &  99.8 &  100 &  98.9 &  95.0 &  93.0 &  98.6 &  90.5 &  85.8 &  89.4 &  2.2 &  14.2 &  99.9 &  99.4 &  99.9 \\
 &\cellcolor{aliceblue}
  cl\_men\_male\_baseline & \cellcolor{aliceblue}  & \cellcolor{aliceblue} 100 & \cellcolor{aliceblue} 100 & \cellcolor{aliceblue} 86.0 & \cellcolor{aliceblue}0.0 & \cellcolor{aliceblue} 99.9 & \cellcolor{aliceblue} 100 & \cellcolor{aliceblue} 99.1 &  \cellcolor{aliceblue} 73.1 & \cellcolor{aliceblue} 49.5 & \cellcolor{aliceblue} 100 &  \cellcolor{aliceblue} 88.1 & \cellcolor{aliceblue} 66.7 & \cellcolor{aliceblue} 100 &  \cellcolor{aliceblue} 43.3 & \cellcolor{aliceblue} 1.3 & \cellcolor{aliceblue} 99.9 &  \cellcolor{aliceblue} 99.9 & \cellcolor{aliceblue} 98.8 \\
 &
  cl\_menself\_male &  95.9 &  99.7 &  100 &  99.6 & 0.0 &  100 &  100 &  99.6 &  85.4 &  74.2 &  99.9 &  98.4 &  92.8 &  100 &  38.2 &  2.5 &  99.8 &  64.2 &  100 \\
 &\cellcolor{aliceblue}
  cl\_men\_female\_baseline & \cellcolor{aliceblue}  & \cellcolor{aliceblue} 100 & \cellcolor{aliceblue} 100 & \cellcolor{aliceblue} 88.8 & \cellcolor{aliceblue} 0.0 & \cellcolor{aliceblue} 99.4 & \cellcolor{aliceblue} 99.5 & \cellcolor{aliceblue} 96.3 &  \cellcolor{aliceblue} 59.7 & \cellcolor{aliceblue} 42.2 & \cellcolor{aliceblue} 99.9 &  \cellcolor{aliceblue} 50.8 & \cellcolor{aliceblue} 39.5 & \cellcolor{aliceblue} 100 & \cellcolor{aliceblue} 4.5 & \cellcolor{aliceblue} 2.9 & \cellcolor{aliceblue} 98.7 & \cellcolor{aliceblue} 99.2 & \cellcolor{aliceblue} 94.3 \\
 &
  cl\_menself\_female &  97.3 &  99.8 &  100 &  96.2 & 0.0 &  100 &  99.4 &  95.9 &  56.3 &  47.0 &  98.1 &  91.3 &  89.6 &  92.4 &  0.7 &  8.5 &  99.2 &  92.5 &  99.9 \\\midrule
\multirow{5}{*}{Classifier-noun agreement} &
  dem\_cl\_swap &  99.6 &  99.6 &  99.8 &  52.5 &  85.7 &  99.8 &  99.5 &  92.3 &  94.2 &  92.2 &  99.0 &  92.8 &  94.1 &  98.9 &  78.5 &  81.4 &  63.0 &  57.5 &  98.3 \\
 &
  cl\_simple\_noun &  98.6 &  95.6 &  96.7 &  61.2 &  85.0 &  98.5 &  98.4 &  88 &  96.4 &  96.6 &  95.9 &  94.5 &  95.9 &  92.4 &  77.9 &  90.4 &  50.1 &  53.1 &  96.3 \\
 &
  cl\_adj\_simple\_noun &  93.2 &  92.1 &  95.5 &  58.9 &  77.1 &  96.5 &  96.5 &  83.8 &  95.8 &  96.1 &  95.2 &  93.6 &  95.2 &  91.0 &  58.4 &  85.5 &  51.6 &  52.6 &  94.5 \\
 &
  cl\_comp\_noun &
  94.0 & 56.2 & 70.7 & 45.6 & 66.3 & 91.3 & 89.9 & 74.2 & 90.6 & 90.3 & 91.1 & 71.2 & 78.5 & 74.7 & 63.3 & 83.6 & 51.8 & 48.6 & 78.3 \\
 &
  cl\_adj\_comp\_noun &
  96.5 & 56.2 & 65.6 & 45.1 & 59.9 & 90.3 & 90.1 & 72.9 & 92.7 & 92.1 & 90.9 & 83.3 & 87.4 & 80.4 & 62.1 & 80.8 & 46.8 & 53.3 & 77.9 \\\midrule
\multirow{5}{*}{Aspect} &
  past\_tense\_guo &
  99.1 & 83.9 & 85.6 & 79.7 & 72.4 & 95.5 & 91.4 & 60.5 & 88.7 & 87.6 & 92.7 & 79.6 & 81.7 & 54.5 & 48.6 & 71.6 & 58.8 & 45.3 & 85.8 \\
 &
  zai\_guo &  98.2 &  55.6 &  88.2 &  78.6 &  65.4 &  97.9 &  98.2 &  90.3 &  87.3 &  88.9 &  97.7 &  61.3 &  84.5 &  65.7 &  67.3 &  89.5 &  49.6 &  54.2 &  91.0 \\
 &
  past\_tense\_le &  99.1 &  76.2 &  70.7 &  78.8 &  73.9 &  65.2 &  61.4 &  39.5 &  81.0 &  77.3 &  51.6 &  49.5 &  64.6 &  37.4 &  31.2 &  37.9 &  53.9 &  43.0 &  57.7 \\
 &
  zai\_no\_le &
  95.0 &
  17.4 &
  50.8 &
  0.8 &
  16.1 &
  85.2 &
  86.7 &
  72.8 &
  55.2 &
  62.5 &
  75.6 &
  31.9 &
  51.4 &
  44.6 &
  56.2 &
  81.4 &
  53.9 &
  43.0 &
  63.1 \\
 &
  zai\_le\_scope &  96.4 &  28.8 &  64.1 &  68.0 &  51.4 &  76.9 &  70.1 &  79.1 &  69.5 &  75.4 &  53.7 &  48.2 &  62.1 &  22.8 &  45.6 &  45.0 &  60.3 &  68.8 &  60.0 \\\midrule
\multirow{2}{*}{Definiteness effect} &
  demonstrative &  96.8 &  94.1 &  99.3 &  48.3 &  49.3 &  98.2 &  98.2 &  82.4 &  97.4 &  96.5 &  94.9 &  55.1 &  65.4 &  92.5 &  59.8 &  25.5 &  27.8 &  16.1 &  70.4 \\
 &
  every &  96.8 &  99.8 &  99.5 &  92.5 &  87.7 &  94.6 &  92.6 &  65.3 &  95.8 &  95.6 &  82.4 &  71.9 &  80.1 &  95.7 &  84.5 &  72.5 &  0.6 &  1.9 &  92.6 \\\midrule
\multirow{3}{*}{Polarity item} &
  any & 90.5 & 87.6 & 89.9 & 95.9 & 93.6 & 65.8 & 86.3 & 94.9 & 97.4 & 97.6 & 75.6 & 93.2 & 95.0 & 33.8 & 61.8 & 83.7 & 60.0 & 45.2 & 63.8 \\
 & even\_wh & 91.4 & 85.3 & 70.3 & 42.3 & 47.7 & 52.4 & 87.4 & 99.1 & 99.5 & 99.5 & 70.8 & 99.1 & 99.6 & 7.1 & 77.6 & 97.4 & 33.0 & 66.6 & 96.2 \\
 & more\_or\_less & 94.1 & 98.0 & 97.7 & 98.6 & 97.6 & 97.9 & 97.5 & 90.0 & 96.9 & 97.6 & 97.6 & 97.3 & 94.9 & 91.8 & 95.1 & 63.8 & 85.6 & 77.0 & 97.7 \\\midrule
\multirow{2}{*}{Relative clause} &
  resumptive noun &  100 &  50.9 &  4.1 &  82.1 &  16.7 &  25.6 &  15.6 &  98.5 &  5.4 &  4.6 &  12.1 &  7.0 &  3.6 &  0.2 &  56.1 &  26.1 &  0.0 &  0.1 &  39.4 \\
 &
  resumptive pronoun &  98.2 &  93.2 &  85.7 &  18.6 &  11.8 &  42.7 &  60.4 &  80.0 &  32.4 &  21.6 &  54.1 &  80.3 &  93.7 &  26.2 &  28.3 &  74.2 &  5.5 &  36.5 &  90.9 \\\midrule
\multirow{2}{*}{\textit{wh} fronting} & bare\_wh & 100 & 100 & 99.9 & 96.6 & 99.7 & 100 & 100 & 99.7 & 99.7 & 98.9 & 99.6 & 99.6 & 99.7 & 75.6 & 86.1 & 98.4 & 7.0 & 36.6 & 100 \\
 & wh\_as\_modifier & 100 & 100 & 99.4 & 90.7 & 88.8 & 99.5 & 99.5 & 99.4 & 99.8 & 99.8 & 99.9 & 95.2 & 99.0 & 60.0 & 76.1 & 98.8 & 19.2 & 52.8 & 100 \\\midrule
  \multicolumn{3}{l}{Average over 9 phenomena} \textbf{97.1} & 75.5 & 78.0 & 72.9 & 55.1 & 84.8 & 83.4 & 81.8 & 81.3 & 79.6 & 83.0 & 72.2 & 75.5 & 59.5 & 57.2 & 53.8 & 41.2 & 45.0 & 75.0 \\
  \bottomrule
\end{tabular}}
\caption{Eighteen LMs' performance on SLING. The blue marked lines are baselines. The baselines are supposed to have an accuracy of $50$\%, meaning the LMs are gender/number neutral.}
\label{tab:18SLNG}
\end{table*}

\subsection{Statistic Tests}\label{appen.analyses}

The results are reported in \tableref{tab:LM.pairs} to \tableref{tab:discussion.cl.noun}.

\begin{table}[!h]
\centering
\resizebox{\linewidth}{!}{%
\begin{tabular}{@{}lllll@{}}
\toprule
LM1 & LM2                   & two-tailed & greater & lesser  \\\midrule
lstm & gpt2-zh              & $0.617$    & --------& --------\\
pert-base & pert-large      & $0.009$**  & $0.005$** & $0.996$ \\
mengzi-base & mengzi-fin    & $0.004$**  & $0.002$** & $0.998$ \\
xlm-R-base & xlm-R-large        & $0.913$    & --------& --------\\
mt5-small &mt5-large        & $0.293$    & --------& --------\\
byt5-small & byt5-large     & $0.277$    & --------& --------\\\bottomrule
\end{tabular}}
\caption{The \textit{p} values of the Wilcoxon signed rank tests of LM pairs.}
\label{tab:LM.pairs}
\end{table}

\begin{table}[!h]
\centering
\resizebox{\linewidth}{!}{%
\begin{tabular}{@{}llll@{}}
\toprule
data & two-tailed & greater    & lesser  \\ \midrule
simple \& simple w/ adj.        & 0.000***    & 0.000*** & 1.000 \\\midrule\midrule
compound \& comp. w/ adj. & 1     & --------   & -------- \\ \bottomrule           
\end{tabular}}
\caption{The results of the Wilcoxon signed rank tests of the simple noun with/withouth a long adjective and the ones with compound nouns.}
\label{tab:discussion.cl.noun.distance}
\end{table}

\begin{table}[!h]
\centering
\resizebox{\linewidth}{!}{%
\begin{tabular}{@{}lllllll@{}}
\toprule
data        & min.  & median    & mean      & max.  & SD        & \textit{p} value   \\ \midrule
male self   & $7.6$ & $84.25$   & $70$      & $100$ & $31.27$   & \multirow{2}{*}{$0.002$**} \\
male pp     & $13.9$& $40.6$    & $52.52$   & $99.9$& $29.42$   & \\ \midrule\midrule
female self & $44$  & $91.65$   & $82.44$   & $99.7$& $19.54$   & \multirow{2}{*}{$0.008$**} \\
female pp   & $18.9$& $70.8$    & $66.36$   & $98.6$& $26.85$   & \\ \bottomrule
\end{tabular}}
\caption{Descriptive statistics of the anaphor (fe)male self and (fe)male self with PP paradigms. The \textit{p} values are from the Wilcoxon signed rank tests.}
\label{tab:discussion.anaphor.male}
\end{table}

\begin{table}[!h]
\centering
\resizebox{\linewidth}{!}{%
\begin{tabular}{@{}lllllll@{}}
\toprule
data        & min.  & median    & mean      & max.   & SD        & \textit{p} value   \\ \midrule
simple      & $50.1$& $95.05$   & $86.83$   & $98.5$ & $15.75$   & \multirow{2}{*}{$0.000$***} \\
compound    &$45.58$& $74.45$   & $73.12$   & $91.26$ & $15.26$   & \\ \midrule\midrule
simple w/ adj. & $51.6$  & $92.85$   & $83.88$   & $96.5$& $16.57$   & \multirow{2}{*}{$0.000$***} \\
comp. w/ adj.  & $45.11$ & $79.13$   & $73.77$   & $92.65$& $16.43$   & \\ \bottomrule
\end{tabular}}
\caption{Descriptive statistics of the classifier \& (adj.) simple noun and compound noun paradigms. The \textit{p} values are from the Wilcoxon signed rank tests.}
\label{tab:discussion.cl.noun}
\end{table}



\begin{figure*}[!h]
\centering
\resizebox{\textwidth}{!}{%
\includegraphics{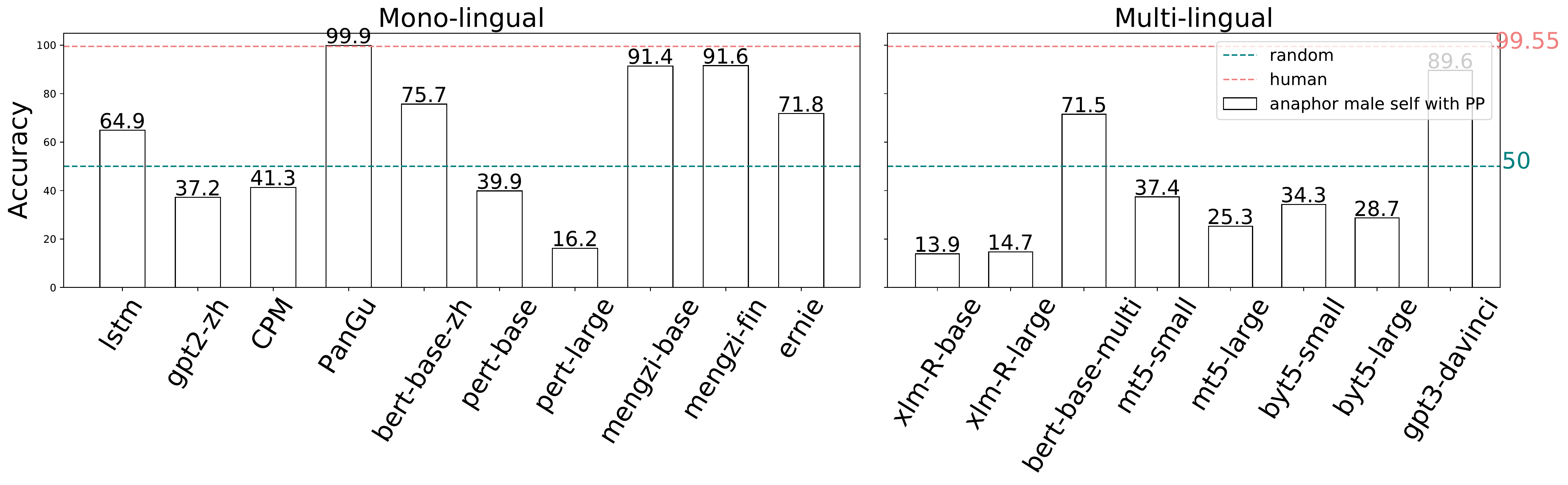}%
}
\caption{The LM accuracy on the anaphor male self with PP paradigm.}
\end{figure*}


\begin{figure*}[!h]
\centering
\resizebox{\textwidth}{!}{%
\includegraphics{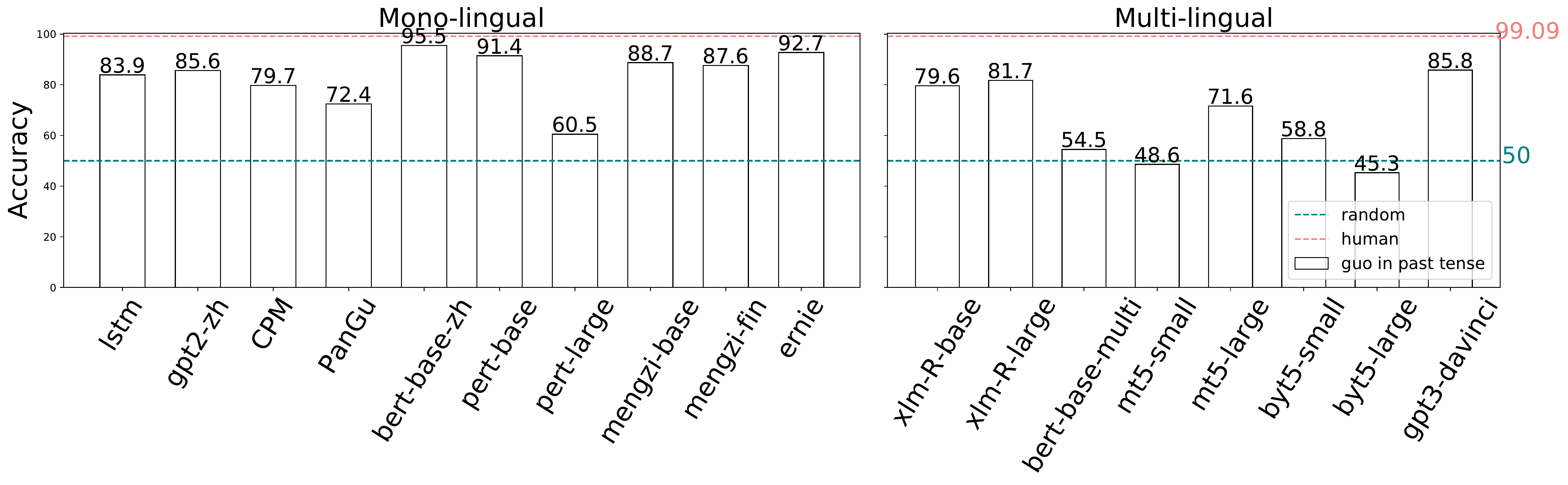}%
}
\caption{The LM accuracy on the guo in past tense paradigm.}
\end{figure*}

\begin{figure*}[!h]
\centering
\resizebox{\textwidth}{!}{%
\includegraphics{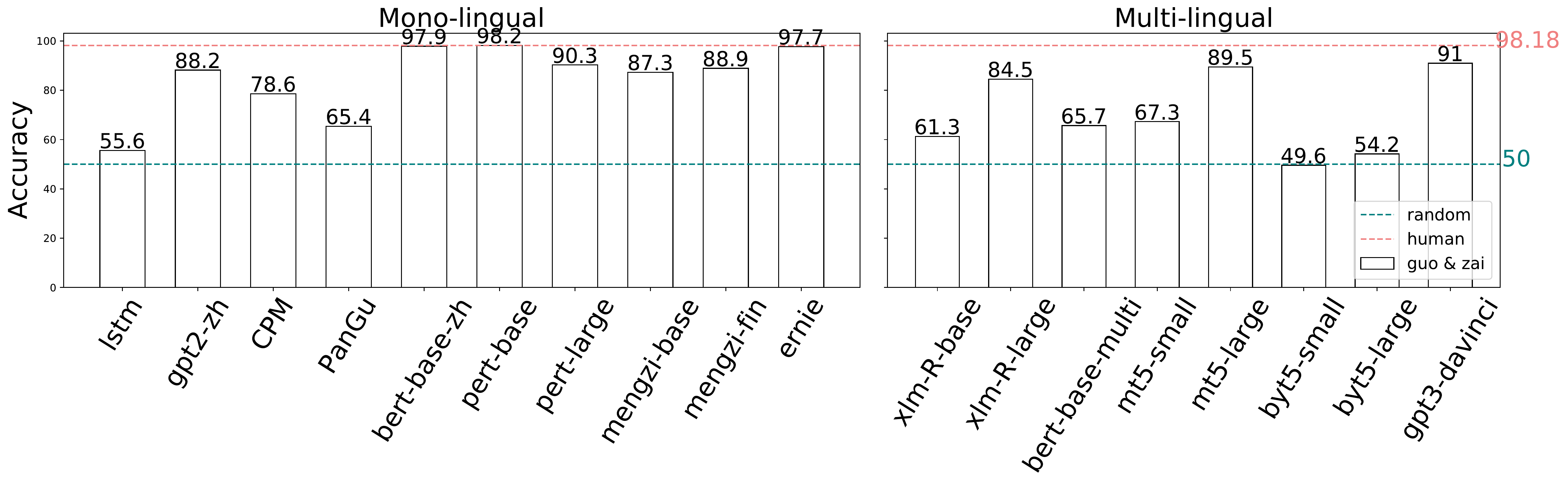}%
}
\caption{The LM accuracy on the guo \& zai paradigm.}
\end{figure*}

\begin{figure*}[!h]
\centering
\resizebox{\textwidth}{!}{%
\includegraphics{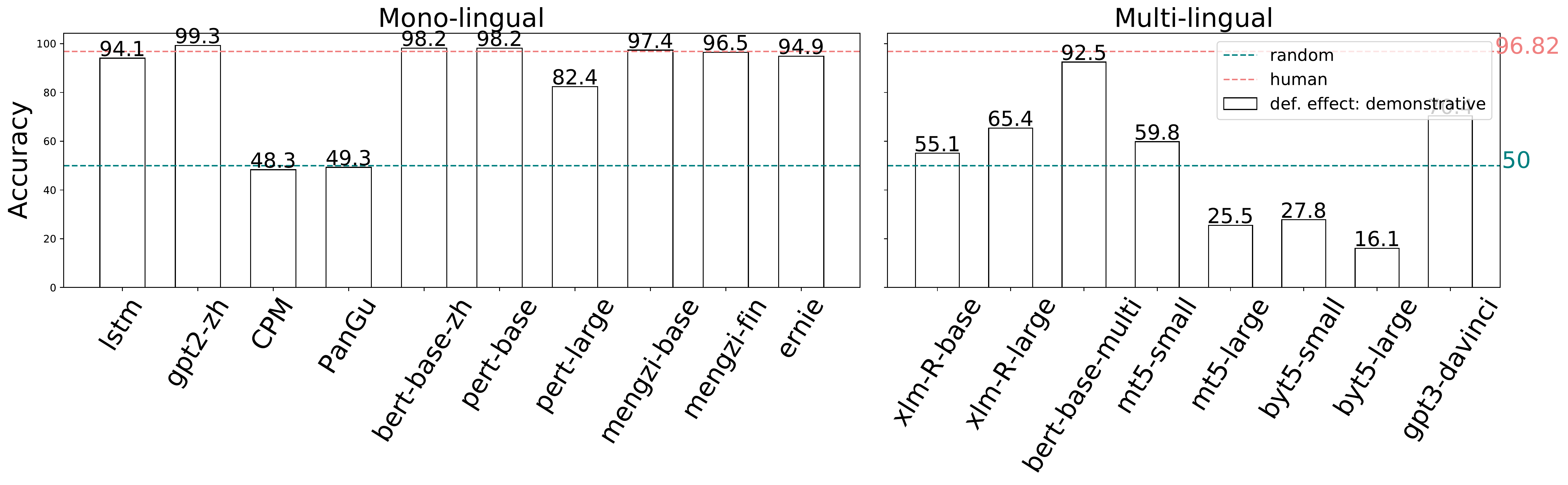}%
}
\caption{The LM accuracy on the definiteness effect with demonstrative paradigm.}
\end{figure*}

\begin{figure*}[!h]
\centering
\resizebox{\textwidth}{!}{%
\includegraphics{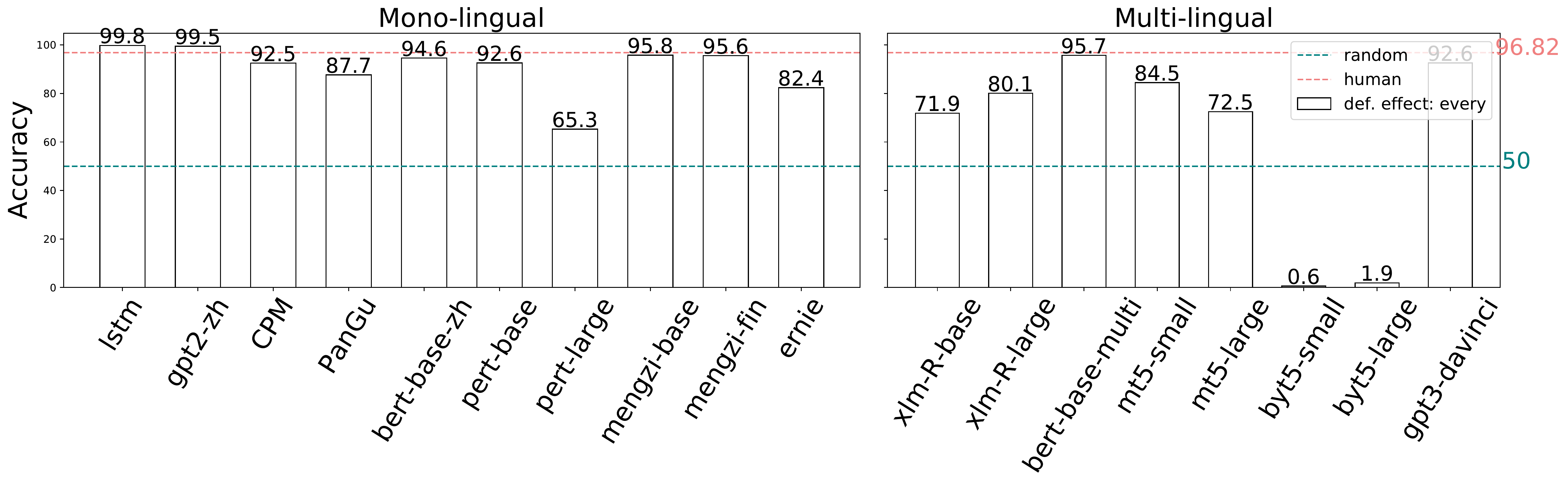}%
}
\caption{The LM accuracy on the definiteness effect with every paradigm.}
\end{figure*}

\end{document}